%% file: ms.tex
\documentclass{article}
\usepackage{listings}
\usepackage{xcolor}
\PassOptionsToPackage{numbers, compress}{natbib}
\usepackage[preprint]{neurips_2026}

\usepackage[utf8]{inputenc} 
\usepackage[T1]{fontenc}    
\usepackage{hyperref}       
\usepackage{url}            
\usepackage{booktabs}       
\usepackage{amsfonts}       
\usepackage{amsmath}        
\usepackage{nicefrac}       
\usepackage{microtype}      
\usepackage{xcolor}         
\usepackage{graphicx}       
\usepackage{subcaption}     
\usepackage{placeins}       
\usepackage{float}

\title{AcceRL: A Distributed Asynchronous Reinforcement Learning and World Model Framework for Vision-Language-Action Models}

\author{%
  \parbox{\textwidth}{\centering
    Chengxuan Lu\thanks{Equal contribution.}, Shukuan Wang\footnotemark[1], 
    Yanjie Li, Yingying Fang, Huoyan Wang, Tian Zhang, Wei Liu, Shiji Jin, Fuyuan Qian, Peiming Li, Chao Xu, 
    Baigui Sun\thanks{Corresponding author.}, Yang Liu\footnotemark[2]
  } \\[2mm]
  IROOTECH TECHNOLOGY \\
  Wolf 1069 b Lab, Sany Group \\[1mm]
  \texttt{\{chengxuan.lu, shukuan.wang, yanjie.li, yingying.fang,} \\
  \texttt{huoyan.wang, tian.zhang, wei.liu, shiji.jin, fuyuan.qian,} \\
  \texttt{peiming.li, chao.xu, baigui.sun, yang.liu1\}@irootech.com}
}

\begin{document}

\maketitle

\input{abstract}

\input{introduction}

\input{related_work}

\input{AcceRL}
\input{AcceRL_WM}
\input{challenge}

\input{experiment}

\input{conclusion}

{
\small
\bibliographystyle{unsrt}
\bibliography{references}
}


\appendix
\input{appendix}





\end{document}

%% file: abstract.tex
\begin{abstract}

Reinforcement learning (RL) for large-scale Vision-Language-Action (VLA) models is severely bottlenecked by synchronization barriers and the high cost of environment data acquisition. To overcome these challenges, we propose AcceRL, a distributed asynchronous RL framework that physically isolates environment rollouts, model inference, and gradient updates. By eliminating the cascading long-tail idle bubbles inherent in synchronous systems, AcceRL maximizes hardware utilization and ensures scalable throughput. Furthermore, AcceRL features a modular design that supports the integration of diverse, plug-and-play world models into its distributed pipeline. Extensive experiments demonstrate that the base framework achieves highly competitive performance across all four LIBERO~\cite{liu2023libero} task suites. Systematically, the asynchronous architecture delivers a $2.4\times$ throughput speedup over leading synchronous baselines. Algorithmically, by leveraging a world model pre-trained on 1,000 offline trajectories, AcceRL achieves up to a $200\times$ improvement in online sample efficiency on LIBERO-Spatial, establishing a robust framework that is both sample-efficient and time-efficient for embodied AI. Code is included in the supplementary material. Code is available at \url{https://github.com/distanceLu/AcceRL}. 


\end{abstract}

%% file: introduction.tex
\section{Introduction}

Vision-Language-Action (VLA) models~\cite{brohan2023rt2, kim2024openvla, octo2023} have rapidly become the default substrate for general-purpose embodied agents, executing complex tasks directly from natural-language instructions. While imitation learning~\cite{zhao2023aloha, chi2023diffusionpolicy} has driven much of the recent progress, it remains limited by poor generalization and error accumulation under covariate shift. The community has therefore turned to reinforcement learning (RL) to optimize long-horizon policies and perform online fine-tuning of pre-trained VLA backbones. Effective RL on VLA models, however, requires massive parallelization to generate the interaction trajectories needed for stable learning, posing an urgent need for an efficient, large-scale RL framework for VLA.

Developing such a framework faces severe challenges in two fundamentally different ways. First, on the efficiency axis, large-scale distributed RL training for VLA models is highly system-intensive. Simulator iteration, multimodal model inference, and parameter updates compete for the same GPU resources, and any pacing mismatch among them causes pipeline stalls. Most existing synchronous frameworks~\cite{liang2018rllib, moritz2018ray} suffer from severe long-tail phenomena that compound this problem. Recent asynchronous frameworks for VLA-RL (e.g., RL-VLA\textsuperscript{3} ~\cite{guan2026rlvla3}) make substantial progress, but their efficiency gains heavily rely on vectorized simulators. While effective on highly parallel benchmarks, this premise breaks down in real-world physical robot deployments, simulators resisting parallelization, or mixed training across heterogeneous tasks. In such regimes, where natural batchability is absent, they struggle to hide environment latencies, leading to severe idle bubbles and a significant drop in throughput. Second, on the data axis, large VLA models are extremely data-hungry. Scaling RL to real-world physical robots faces severe data-collection bottlenecks. Introducing a learned World Model (WM) to synthesize experience naturally resolves this physical sampling limit, paving the way for real-world adaptability. Yet integrating it into a high-performance distributed RL pipeline remains unsolved. Existing WMs (like the Dreamer family~\cite{hafner2023dreamerv3}) tightly couple policy and dynamics within shared latent spaces, severely hindering the plug-and-play integration of independent pre-trained models. Conversely, pixel-level video WMs~\cite{bruce2024genie, alonso2024diffusion} provide a naturally decoupled visual interface but have only served as standalone simulators, never fully integrated into an asynchronous RL loop.

To bridge this gap, we introduce AcceRL, a modular and fully asynchronous framework tailored for large-scale VLA models. AcceRL achieves complete system decoupling by physically isolating training, inference, and sampling into independent asynchronous streams that communicate through shared buffers, without any batchability or vectorization assumptions on the producer side. Furthermore, it pioneers the integration of a pixel-level trainable world model as a co-equal experience source alongside real-environment workers, bridged by a value recomputation mechanism. Our key contributions are as follows:
\begin{itemize}
\item \textbf{High-throughput asynchronous RL architecture:} AcceRL realizes full asynchrony among rollout, inference, and training. By removing synchronization barriers, eliminating frequent mode-switching, and mitigating long-tail latencies, it maximizes hardware utilization. Crucially, unlike prior asynchronous frameworks dependent on vectorized simulators, AcceRL sustains consistently high throughput and resource efficiency even in complex environments where natural batchability is absent, achieving a $2.4\times$ throughput speedup over leading synchronous baselines.
\item \textbf{Pioneering integration of a trainable world model:} AcceRL pioneers the integration of a pluggable, pixel-level trainable world model into a distributed asynchronous RL pipeline. The world model operates through a pixel-space interface, which helps preserve the simulator’s observation-action structure and facilitates compatibility with different VLA backbones. To absorb the compound staleness arising from full asynchrony, we combine an optimized value recomputation mechanism with the adopted GIPO~\cite{lu2026gipo} algorithm. By leveraging a world model pre-trained on 1,000 offline trajectories (which are out-of-distribution and more cost-effective to collect than online samples) and breaking through physical sampling limitations via imagination learning, AcceRL improves online sample efficiency by up to $200\times$ over a strong asynchronous model-free baseline on LIBERO-Spatial~\cite{liu2023libero} under matched compute.
\item \textbf{Empirical effectiveness and competitive performance}: AcceRL achieves highly competitive performance on the LIBERO benchmark, attaining the best results among the compared methods on most task suites while demonstrating outstanding system scalability, high computational efficiency, and rapid convergence with high data efficiency.
\end{itemize}

%% file: related_work.tex
\section{Related work}
\label{sec:related_work}

\subsection{Evolution of Distributed Reinforcement Learning Frameworks}
\label{subsec:evolution_frameworks}
Historically, scaling RL relies on synchronous distributed frameworks. For Large Language Models (LLMs), frameworks like OpenRLHF~\cite{hu2024openrlhf} and DeepSpeed Chat~\cite{yao2023deepspeed} combine high-performance inference engines with distributed training backends to optimize memory during synchronous parameter updates. Similarly, the veRL framework~\cite{sheng2025hybridflow} adopts a hybrid flow architecture that separates control logic from tensor operations while maintaining lockstep synchronization between generation and training.  This paradigm naturally extends to the VLA domain. Systems such as siiRL~\cite{wang2025distflow} and RLLAVA~\cite{zhao2025rllava} use colocated execution with synchronized actor-learner resource sharing. RLinf-VLA~\cite{zang2026rlinfvlaunifiedefficientframework} further optimizes this with a hybrid fine-grained pipeline to overlap rendering and training, but retains the core synchronous stepping mechanism. Despite mathematical stability, synchronous architectures suffer from long-tail effects: the central learner waits for all distributed Rollout Workers to finish trajectory collection before gradient updates~\cite{cipar2013solving}, which maintains the bounded policy lag required by PPO~\cite{schulman2017proximal}. LLM post-training has two-level long-tail latency from episode-level token generation and cluster-level from network communication. Moreover, VLA frameworks have a third more complex long tail. Because physical simulators require dynamic computational overhead, the entire distributed system is forced to idle while waiting for the slowest physics step.

To mitigate synchronization overhead and pipeline bubbles, the field has shifted to asynchronous/decoupled architectures. For LLMs, frameworks decouple generation and training: Laminar ~\cite{tong2025laminar} uses trajectory-level asynchrony for long-tail text generation. AReaL~\cite{fu2026areallargescaleasynchronousreinforcement} fully decouples them to reduce GPU underutilization from reasoning tokens. DistFlow~\cite{wang2025distflow}, LlamaRL~\cite{wu2025llamarl} and ROLL~\cite{wang2025reinforcement} also adopt decoupled/asynchronous strategies, but these are tailored for in-memory closed-loop autoregressive generation, misaligned with embodied agents' external dependencies. RL-VLA\textsuperscript{3} ~\cite{guan2026rlvla3} proposes triple-level decoupling for embodied AI but is not fully asynchronous, failing in complex physical scenarios.  To overcome this physical wall-clock bottleneck, the system needs to achieve complete physical-logical decoupling. AcceRL accomplishes this fully asynchronous model by entirely separating the training, inference, and rollout.

\subsection{World Models and Model-Based RL}
\label{subsec:world_models}
Model Based Reinforcement Learning fundamentally accelerates convergence by learning a predictive model of environment dynamics to substitute costly physical interactions with computational predictions. A landmark in this field is the Dreamer series~\cite{hafner2023dreamerv3} which projects high dimensional observations into a compact discrete latent space. By optimizing policies within this differentiable dream using Recurrent State Space Models can achieve high sample efficiency in environments with dense rewards such as Atari games.

While latent space models are efficient for coarse grained tasks transitioning to pixel level world models based on Diffusion Models~\cite{ho2020denoising} becomes a structural necessity for embodied VLA models. Pixel-level prediction retains fine-grained visual context critical for learning complex contact dynamics. Using raw pixels as a universal interface enables policy-world model decoupling for independent scaling and integration, free of shared latent encoder constraints. Additionally, diffusion models avoid complex joint loss optimization in heterogeneous model grafting, offering a stable, high-fidelity virtual sandbox for fine-tuning pre-trained models.
Building on these pixel level predictions frameworks such as Genie~\cite{bruce2024genie} and DIAMOND~\cite{alonso2024diffusion} have demonstrated the potential of world models as infinite risk free simulators. However integrating these models into VLA training loops introduces severe VRAM bottlenecks. Recent works like WoVR~\cite{jiang2026wovr} attempt to bridge this gap but rely on a collocated GPU allocation strategy. This necessitates frequent offloading and onloading of massive parameters between CPU and GPU leading to significant compute idleness and low throughput.By physically isolating policy and world models on dedicated hardware AcceRL eliminates memory swapping and reduces communication to trajectory passing, which supports independent dynamics control and resource reuse.

%% file: AcceRL.tex
\section{AcceRL: A Fully Asynchronous Training Framework}

\begin{figure}[htbp]
    \centering
    \includegraphics[width=\textwidth]{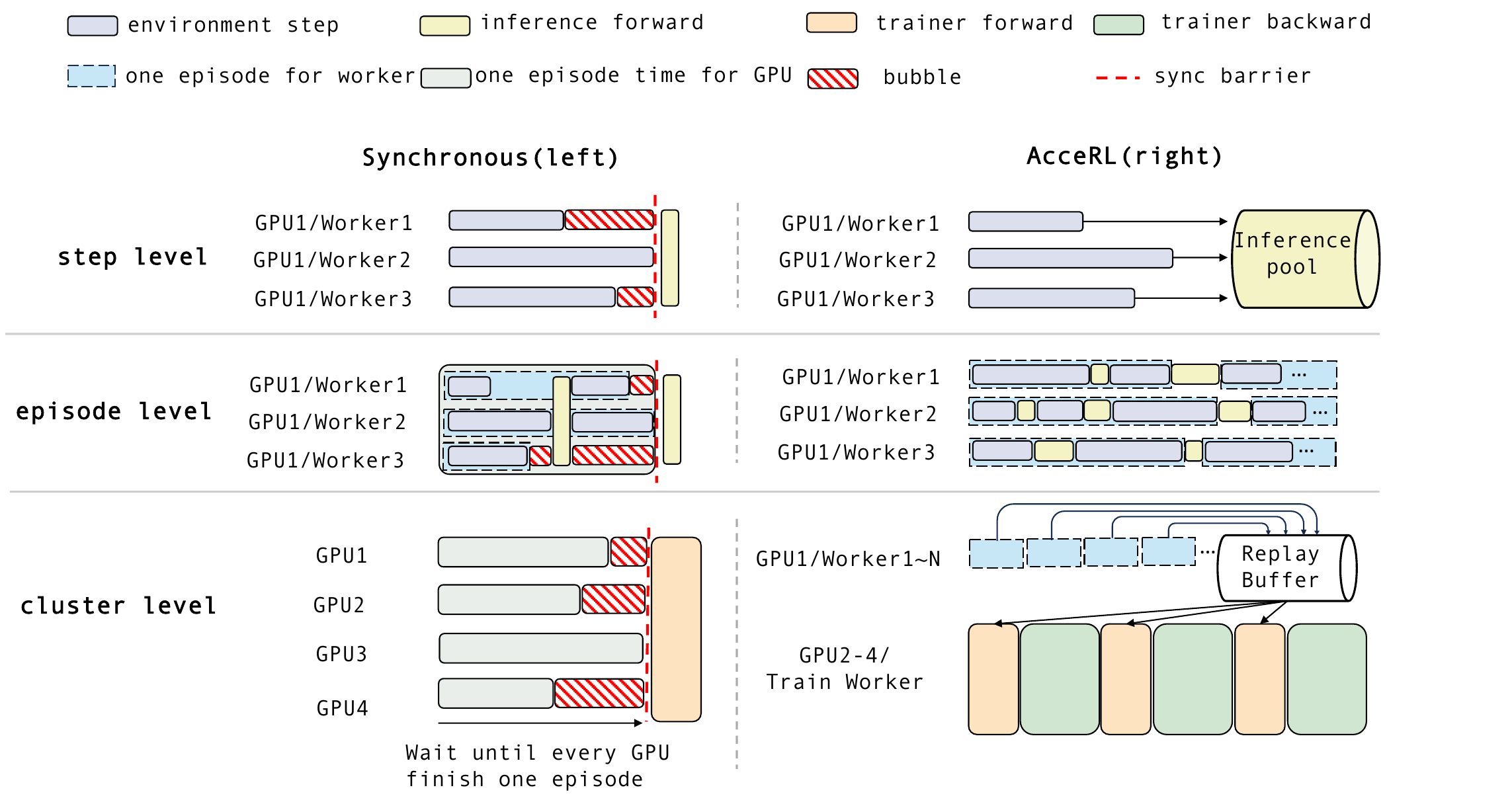}
    \caption{\textbf{Timelines and GPU bubbles across three levels of synchronous RL framework (left) and the asynchronous framework AcceRL (right).} The synchronous framework suffers from long-tail effects at the step, episode, and cluster levels, resulting in severe forced waiting bubbles (red hatched areas). By introducing a dual asynchronous mechanism, AcceRL substantially reduces synchronization barriers and long-tail idle periods, enabling more efficient computation.}
    \label{fig:bubble_detail}
\end{figure}

As shown in Figure~\ref{fig:bubble_detail}, traditional synchronous RL frameworks suffer from heavy synchronization overhead caused by a cascading long-tail effect across three distinct levels:

\begin{itemize}
\item Step-Level Long-Tail: Physical simulation times vary across environment instances. Synchronous setups force the GPU to wait for the slowest worker to finish a step before executing batched inference, causing step-level idle bubbles.
\item Episode-Level Long-Tail: Episodes end at different times due to varying termination conditions. Synchronous pipelines prevent the GPU from starting new episodes until all parallel workers finish, forcing early-finishing environments into idleness.
\item Cluster-Level Long-Tail: Global synchronization barriers require all GPUs to complete their rollouts before executing a gradient update. This caps the entire cluster's throughput at the slowest GPU, creating massive system-wide bubbles.
\end{itemize}

As training scales to multi-billion parameter VLA models across large clusters, these three layers of latency compound exponentially, leading to prohibitive hardware underutilization.

To overcome these bottlenecks, we propose AcceRL (Figure~\ref{fig:accerl}), a high-throughput asynchronous framework designed for large-scale Embodied AI. The core advantage of AcceRL is the complete decoupling of training, rollout, and inference for asynchronous parallel execution. It utilizes the NCCL weight synchronization mechanism, an non-blocking FIFO replay buffer mechanism, and a data prefetching mechanism to improve training efficiency (see the appendix~\ref{sec:refinements} for details). 
\subsection{Macro-Asynchrony: Decoupling Training and Rollout}

\begin{figure}[htbp]
    \centering
    \begin{minipage}[t]{0.34\textwidth}
        \centering
        \includegraphics[width=\linewidth]{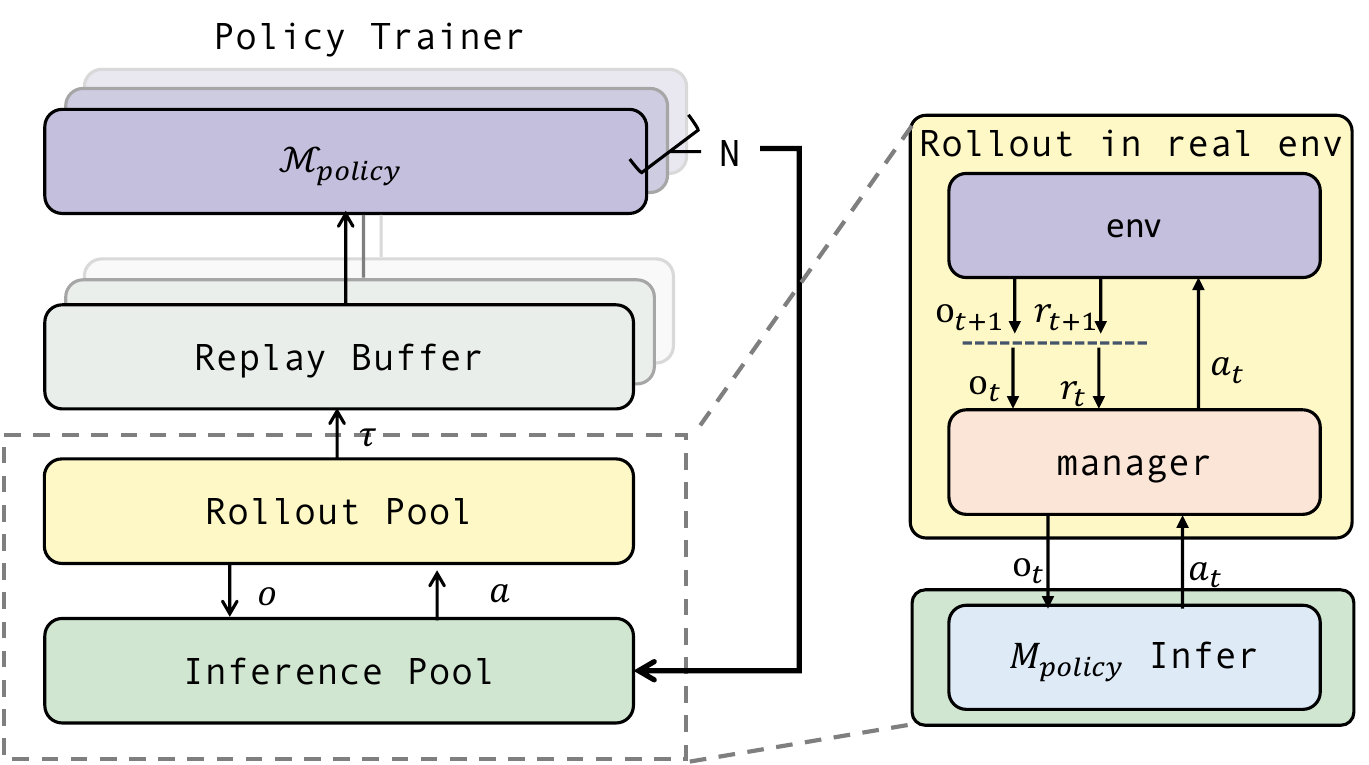}
        \subcaption{AcceRL}
        \label{fig:accerl}
    \end{minipage}\hfill
    \begin{minipage}[t]{0.63\textwidth}
        \centering
        \includegraphics[width=\linewidth]{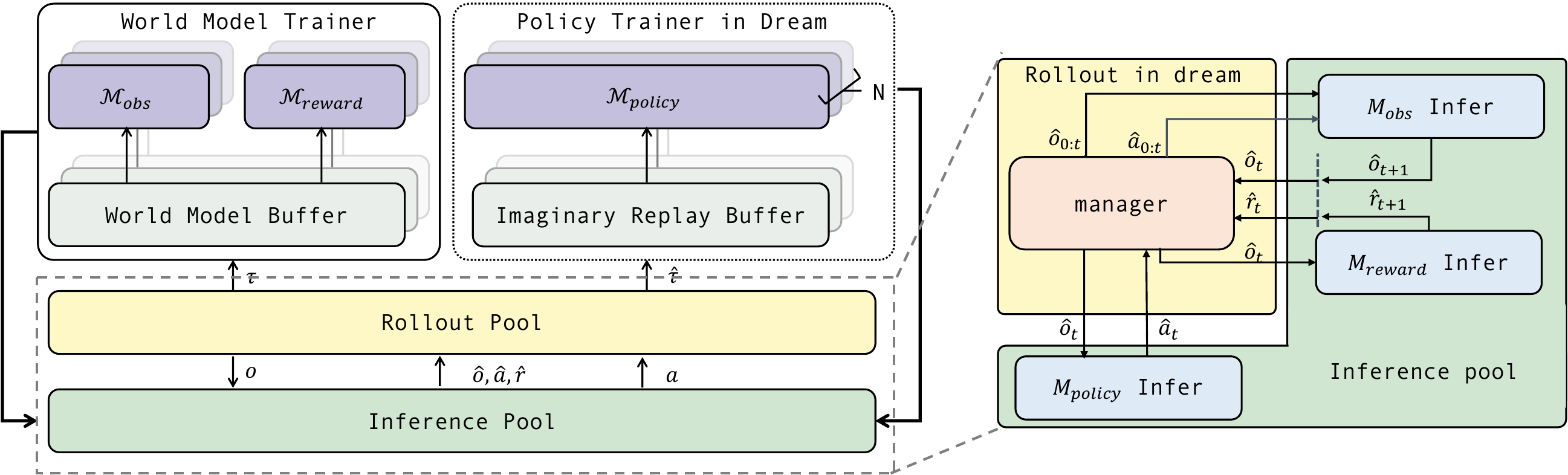}
        \subcaption{AcceRL-WM}
        \label{fig:accerl_wm}
    \end{minipage}
    \caption{\textbf{The pipeline of AcceRL base backbone (left) and world-model-augmented mode (right).} The framework isolates training, inference, and rollout into independent asynchronous streams. By integrating a plug-and-play world model, AcceRL bypasses physical simulation constraints through imagination rollouts to significantly enhance sample efficiency.}
\end{figure}

AcceRL eliminates cluster-level synchronization overhead by fully decoupling Rollout and Trainer Workers into a non-blocking circular pipeline, bypassing the strict lockstep or bounded staleness of prior frameworks~\cite{fu2026areallargescaleasynchronousreinforcement}. As illustrated in Figure~\ref{fig:accerl}, Rollout Workers execute interruptible rollouts, streaming trajectory segments immediately to a FIFO distributed replay buffer $\mathcal{B}$. Concurrently, Trainer Workers sample batches from $\mathcal{B}$ for continuous parameter updates. To address the resulting policy lag from stale data without sacrificing throughput, we utilize value recomputation and GIPO (Section~\ref{sec:policy lag}).


For large-scale efficiency, Trainer Workers employ ZeRO-2~\cite{rajbhandari2020zero} to partition optimizer states and gradients, supporting larger micro-batch sizes. Updated parameters are subsequently synchronized to Inference Workers via NCCL broadcast. This architecture ensures maximum-speed training and significantly reduces end-to-end wall-clock time.

\subsection{Micro-Asynchrony: Decoupling Interaction and Inference}
While macro-asynchrony removes global barriers, the second level of asynchrony eliminates micro-level idle bubbles caused by step- and episode-level long-tails. Traditional tightly coupled simulation and inference suffer from severe straggler effects and poor GPU batching due to highly variable environment step times and episode horizons. To overcome this problem, AcceRL decouples physical interaction from policy inference via an Inference-as-a-Service paradigm.

Rollout Workers manage environment instances. After generating an observation $o_t$, they send asynchronous requests to a centralized inference pool and then suspend immediately to avoid local computation delays. Every Inference Worker maintains a request queue $Q$ and trigger batched forward passes using a dynamic window mechanism to balance GPU utilization and latency:
\begin{equation}
\text{Trigger} = (|Q| \ge B) \lor (t_{now} - t_{first} \ge T_{max})
\end{equation}
where $B$ is the target batch size, $t_{now}$ is the current timestamp, and $T_{max}$ is the maximum allowable wait time since the oldest request's arrival $t_{first}$. Computed actions are then returned to the respective workers.

After each episode terminates, completed sequences are packaged as trajectories $\tau$ and stored in the replay buffer $\mathcal{B}$:
\begin{equation}
\tau = \left( o_{1:T+1}, a_{1:T}, r_{1:T}, \mu_{1:T}, v_{1:T}, \tilde{v}_{T+1}, \text{done} \right),
\end{equation}
where $o_t$, $a_t$, and $r_t$ denote standard observations, actions, and rewards; $\mu_t$ represents the behavior policy probabilities for off-policy correction; $v_t$ and $\tilde{v}_{T+1}$ are state and bootstrapped value estimates; and $\text{done}$ indicates natural episode termination.

%% file: AcceRL_WM.tex
\section{AcceRL with World Model}



While AcceRL maximizes hardware utilization, direct physical environment interaction inherently bottlenecks sample efficiency. To overcome this, the framework introduces an augmented mode, AcceRL-WM, which integrates a plug-and-play world model to enable rapid, sample-efficient imagination rollouts. This world model comprises an observation model $M_{obs}$ to simulate dynamics and a reward model $M_{reward}$ to evaluate success. As illustrated in Figure \ref{fig:accerl_wm}, AcceRL-WM deploys independent Trainer and Inference Workers for these models and partitions the distributed replay buffer into $\mathcal{B}_{wm}$ for real-world state transitions and $\mathcal{B}_{img}$ for synthetic trajectories. 

\subsection{Alternating Rollout and Learning in Imagination}
As Figure~\ref{fig:accerl_wm} shown, to generate data for the world model and policy, Rollout Workers use an alternating sampling strategy. A real trajectory $\tau$ is first collected from the real-world environment and serves two purposes. It is added to $\mathcal{B}_{wm}$ as training data for the world model, and cached locally to supply initial grounding states for imagination rollouts.

Following the real rollout, the worker generates numerous imagined trajectory episodes, denoted with a hat $\hat{\cdot}$.The imagination pipeline operates as follows. First, a real frame $o_t$ is sampled from the local episode buffer as the initialization and starting state, where $\hat{o}_t = o_t$. Next, the worker invokes $M_{policy}$ to produce the imagined action $\hat{a}_t$. Then, it queries  $M_{obs}$ to generate a predicted next frame $\hat{o}_{t+1}$. Meanwhile, the predicted frames are fed into $M_{reward}$ to estimate the success probability of the current state and output the corresponding termination signal $\hat{done}$.

Crucially, to mitigate the compounding errors inherently caused by long-term autoregressive generation, these imagined trajectories are strictly restricted to a fixed horizon $H$ rather than building full executions. The resulting short-term trajectory episode $\hat{\tau}$ is added to $\mathcal{B}_{img}$:
\begin{equation}
    \hat{\tau} = \left( \hat{o}_{t:t+H+1}, \hat{a}_{t:t+H}, \hat{r}_{t:t+H}, \hat{\mu}_{t:t+H}, \hat{v}_{t:t+H}, \tilde{v}_{t+H+1}, \hat{done} \right)
\end{equation}

For these imagined transitions, AcceRL utilizes a potential-based reward structure formulated as:
\begin{equation}
    \hat{r}_\tau = M_{reward}(\hat{o}_{t+1}) - M_{reward}(\hat{o}_t)
\end{equation}
where $\hat{o}_t$ denotes the synthesized observation at step $t$, which can be a multi-frame stack to capture temporal dynamics. By defining the reward as the marginal difference in success probability between successive synthetic states, this formulation helps preserve policy invariance. It accelerates training by providing dense, informative guidance without altering the fundamental objective of the original manipulation task.

\subsection{Training pipeline of AcceRL-WM}
Rather than relying on a static pre-trained world model, AcceRL-WM continuously fine-tune both $M_{obs}$ every $T_{obs}$ cycles and $M_{reward}$ every $T_{reward}$ steps. This constant grounding ensures the dream space adapts to novel states explored by the evolving policy.  
During the training phase, the Trainer Worker manages three fully independent concurrent optimization loops with different update frequencies to stabilize $M_{policy}$, $M_{obs}$, and $M_{reward}$.
\begin{itemize}
    \item $M_{policy}$ Optimization: The policy trainer continuously samples imagined data batches $\hat{\tau}$ from the imaginary replay buffer $\mathcal{B}_{img}$ to update $M_{policy}$, completely bypassing the physical simulator's latency.
    \item $M_{obs}$ Optimization: The observation model is fine-tuned every $T_{obs}$ cycles using the real-world trajectory data sampled from $\mathcal{B}_{wm}$, ensuring the dream space remains visually and physically accurate.
    \item $M_{reward}$ Optimization: The reward model performs regression on real $(o_t, r_t)$ pairs sampled from $\mathcal{B}_{wm}$ every $T_{reward}$ steps to refine its success probability estimations.
\end{itemize}
To minimize communication overhead, updated weights are broadcasted to their respective nodes in the Inference Pool only when an actual update occurs. 

%% file: challenge.tex
\section{Mitigating Policy Lag in Asynchronous Training}
\label{sec:policy lag}
While AcceRL's fully asynchronous and decoupled architecture maximizes hardware utilization, it introduces policy lag.
The macro-asynchrony between the Rollout Workers and Trainer Workers causes the behavior policy $\mu$ used during data collection to become stale compared to the continuously updating learner policy $\pi$. This staleness introduces off-policy bias. To mitigate the off-policy bias induced by policy lag, the Trainer Worker employs a dual mechanism: it utilizes value recomputation for real-time advantage correction to eliminate staleness in value estimation, and applies GIPO~\cite{lu2026gipo} for robust gradient calibration to stabilize policy updates against delayed trajectories.

\textbf{Value Recomputation}:
In fully asynchronous RL frameworks, policy lag introduces a certain degree of off-policy bias, necessitating state value recomputation to realign advantage estimation with the current policy. Traditionally, this requires an additional full forward pass over the dataset, severely bottlenecking throughput. Furthermore, global random data shuffling and synchronous aggregation for advantage statistics introduce memory fragmentation and communication latency.

To resolve these bottlenecks, we propose a low-overhead pipeline. Instead of a separate preprocessing phase, we push Generalized Advantage Estimation (GAE)~\cite{schulman2015high} directly to the post-forward-propagation stage of micro-batch training. Given our single-epoch update design, these on-the-fly advantage estimates closely approximate those obtained by forced re-inference, enabling timely value correction while avoiding the substantial cost of additional policy forward passes. At the data level, we replace global shuffling with sequential micro-batch slicing, ensuring contiguous memory access. Finally, we mask communication latency via asynchronous lag normalization: the system uses the previous step's global moving statistics for current normalization, deferring synchronous aggregation to the end of backpropagation.

By eliminating redundant inference, streamlining memory access, and deeply overlapping communication with computation, these optimizations achieve a 30\% increase in end-to-end training speed. They effectively mitigate policy lag without compromising convergence. Mathematical equivalence proofs and implementation details is provided in Appendix~\ref{sec:value_recomputation_suppl}.

\textbf{Gradient Calibration (GIPO)}:
Even with updated advantages, extreme policy lag can cause massive divergence if standard PPO is used, as its hard clipping mechanism zeroes out gradients for highly stale data, stalling the learning process. To protect the policy, AcceRL employs Gaussian Importance sampling Policy Optimization (GIPO)~\cite{lu2026gipo}. Let $\rho_t(\theta) = \pi_\theta(a_t \mid o_t) / \mu(a_t \mid o_t)$ denote the importance ratio between the learner and behavior policies, and $\bar{\rho}_t$ be its stop-gradient version. GIPO introduces a Gaussian trust weight, governed by hyperparameter $\sigma$, to softly damp extreme ratios:
\begin{equation}
    \omega(\bar{\rho}_t; \sigma) = \exp \left( -\frac{1}{2} \left( \frac{\log(\bar{\rho}_t)}{\sigma} \right)^2 \right)
\end{equation}
\begin{equation}
    \mathcal{L}^{\text{GIPO}}_\pi (\theta) = -\mathbb{E}_{(o_t, a_t) \sim \mathcal{B}} \Big[ \omega(\bar{\rho}_t; \sigma) \cdot \rho_t(\theta) \cdot A_t \Big]
\end{equation}
where $A_t$ represents the advantage estimate. This smooth, continuous penalty inherently bounds the policy update magnitude without abruptly discarding useful gradients, allowing the framework to safely extract learning signals from highly delayed trajectories.

%% file: experiment.tex
\section{Experiments}
We evaluate our framework across two diverse simulated domains: the comprehensive LIBERO benchmark~\cite{liu2023libero} (encompassing the Spatial, Object, Long, and Goal suites) and contact-rich continuous control tasks in ManiSkill~\cite{DBLP:conf/nips/MuLXYLLTHJ021}.

\subsection{Throughput Comparison}
\begin{figure*}[htbp]
    \centering
    \begin{subfigure}[t]{0.42\textwidth}
        \centering
        \includegraphics[width=\textwidth]{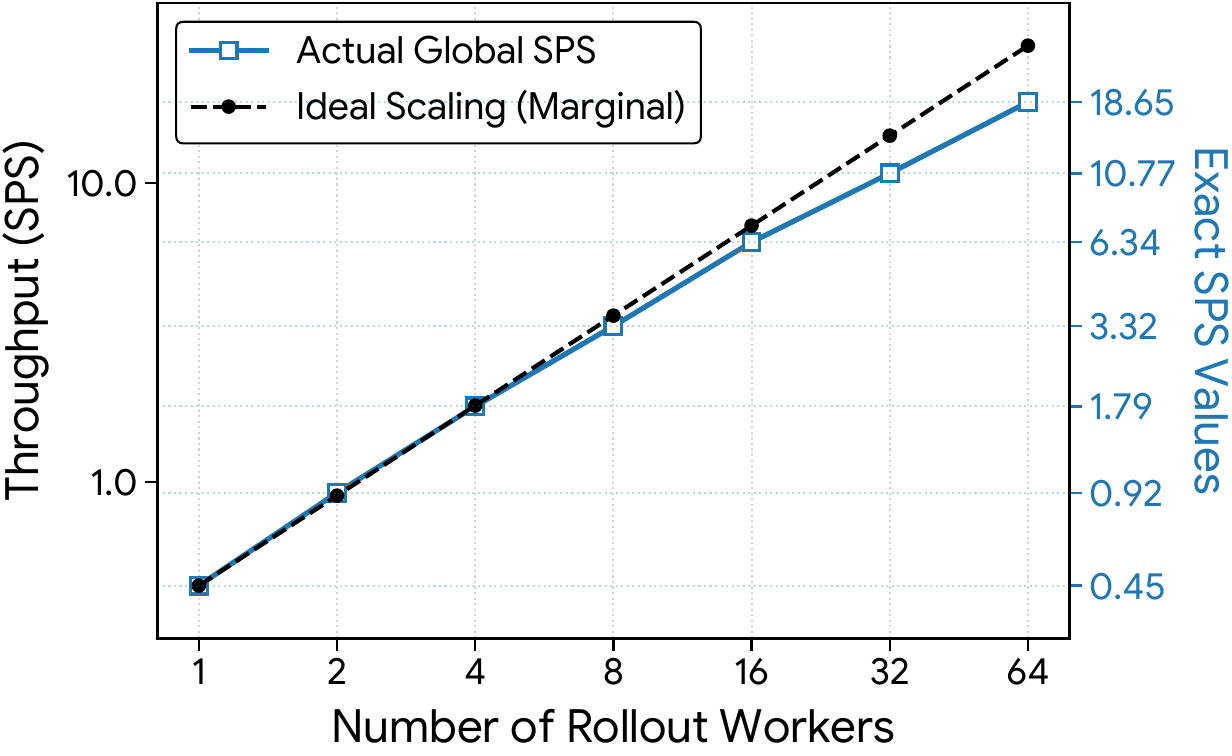}
        \caption{ }
        \label{fig:scale_rollout}
    \end{subfigure}
    \hfill
    \begin{subfigure}[t]{0.42\textwidth}
        \centering
        \includegraphics[width=\textwidth]{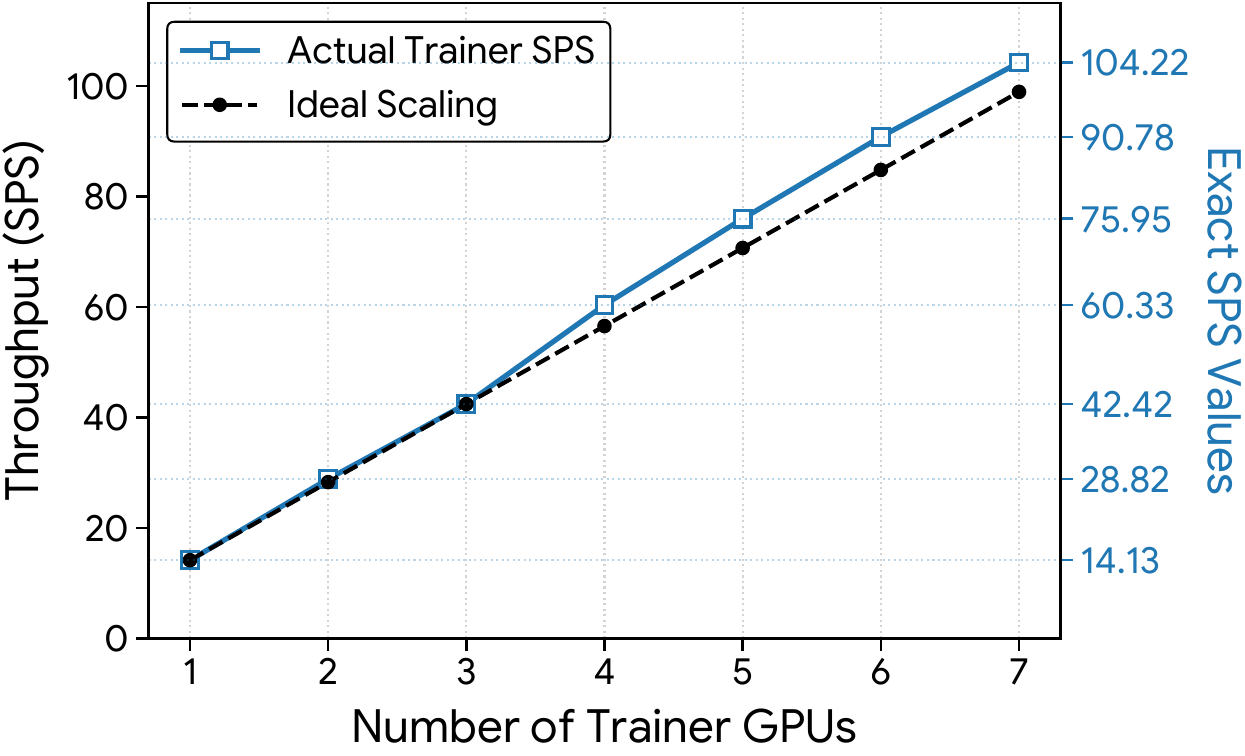}
        \caption{ }
        \label{fig:scale_trainer}
    \end{subfigure}
    \caption{\textbf{Scalability performance of the AcceRL framework with decoupled data labels.} \textbf{(a)} Throughput scaling demonstrates near-linear performance up to 64 Rollout Workers. \textbf{(b)} Trainer scalability shows that the actual trainer SPS closely tracks the ideal marginal scaling curve when scaling up to 7 GPUs. Both configurations effectively maintain high hardware utilization without suffering from severe communication overhead.}
    \label{fig:accerl_scalability}
\end{figure*}
\FloatBarrier
We first evaluate the training throughput and scalability of AcceRL. By decoupling environment interaction from action generation and using centralized dynamic batching, AcceRL effectively reduces GPU bubbles caused by long-tail inference latency. To demonstrate the efficiency of this design, Figure~\ref{fig:accerl_scalability} shows the scaling performance of the system. Notably, when scaling the trainer to 7 GPUs(H200), AcceRL exhibits a super-linear scaling effect, reaching 104.22 SPS (samples per second). This super-linear scaling is primarily driven by DeepSpeed ZeRO-2~\cite{huggingface_deepspeed}. As the cluster expands, it enables the system to accommodate a larger micro-batch size per GPU without out-of-memory errors. The increased batch size heavily amortizes kernel launch overheads and maximizes Tensor Core utilization, allowing the actual throughput to consistently surpass the ideal marginal scaling curve while maintaining a high GPU utilization of over 94\%. Moreover, Table~\ref{tab:system_efficiency} highlights how this near-saturated utilization ($\sim$94--95\%) creates a stark contrast with existing baselines. By circumventing the severe offloading overheads and physical simulator bottlenecks that limit SimpleVLA ($\sim$30--40\% utilization) and RLinf ($\sim$50--60\% utilization), AcceRL delivers a superior and stable sampling throughput of  42.4 SPS. Ultimately, this allows AcceRL to reach 2,000 training steps in just 6.5 hours delivering a 2.4 times  speedup over RLinf and nearly a 2.6 times speedup over the synchronous SimpleVLA baseline.


\begin{table}[htbp]
    \centering
    \caption{\textbf{Comparison of system efficiency and training throughput on 4$\times$ H200 GPUs.}
    Under the same 4$\times$ H200 hardware setting, AcceRL achieves higher trainer GPU utilization and sampling throughput than the compared baselines, substantially reducing the estimated wall-clock time required to reach 2K training steps.}
    \label{tab:system_efficiency}
    \begin{tabular}{lcccc}
        \toprule
        \textbf{Framework} & \textbf{Peak Trainer} & \textbf{Throughput (SPS)} & \textbf{Throughput (SPS)} & \textbf{Est. Time to} \\
         & \textbf{GPU Util.} & \textbf{@ 90\% Succ.} & \textbf{@ 99\% Succ.} & \textbf{2K Steps} \\
        \midrule
        SimpleVLA (VeRL) & $\sim$30--40\% & $\sim$21.4  & $\sim$28.5  & $\sim$16.7 hrs \\
        \addlinespace
        RLinf            & $\sim$50--60\% & $\sim$25.2 & $\sim$35.6 & $\sim$15.8 hrs  \\
        \addlinespace
        AcceRL (Ours)    & \textbf{$\sim$94--95\%} & \textbf{$\sim$42.4} & \textbf{$\sim$42.4} & \textbf{$\sim$6.5 hrs} \\
        \bottomrule
    \end{tabular}
\end{table}




\subsection{Task Performance on LIBERO}
\label{sec:libero_performance}

We evaluate AcceRL against the OpenVLA-OFT supervised baseline~\cite{kim2025fine} and recent state-of-the-art RL frameworks (RLinf-VLA~\cite{zang2026rlinfvlaunifiedefficientframework}, SimpleVLA-RL~\cite{li2025simplevlarlscalingvlatraining}) across four diverse LIBERO task suites.

\begin{table}[htbp]
    \centering
    \caption{\textbf{Task performance comparison on the LIBERO benchmark.} AcceRL achieves superior success rates across most categories. Notably, it attains a 99.1\% success rate on Long tasks, demonstrating stability in long-horizon scenarios where supervised fine-tuning typically falters.}
    \label{tab:libero_performance}
    \begin{tabular}{lcccc}
        \toprule
        \textbf{Method} & \textbf{Spatial (\%)} & \textbf{Object (\%)} & \textbf{Goal (\%)} & \textbf{Long (\%)} \\
        \midrule
        Ours (AcceRL)                                   & \textbf{99.6} & \textbf{100.0} & 98.8          & \textbf{99.1} \\
        SimpleVLA-RL~\cite{li2025simplevlarlscalingvlatraining} & 99.4          & 99.8           & \textbf{99.2} & 98.5          \\
        RLinf-VLA~\cite{zang2026rlinfvlaunifiedefficientframework}                        & 99.4         & 99.8           & 98.8          & 94.0          \\
        OpenVLA-OFT                                     & 96.2          & 98.3           & 96.2          & 90.7          \\
        \bottomrule
    \end{tabular}
\end{table}
\FloatBarrier

As summarized in Table~\ref{tab:libero_performance}, AcceRL achieves the best or tied-best performance on most task suites and remains competitive on Goal. OpenVLA-OFT struggles on LIBERO-Long tasks (90.7\%) due to compounding errors inherent in supervised imitation learning. While recent RL frameworks improve upon this supervised baseline, they still exhibit performance degradation in demanding long-horizon scenarios. In contrast, by formulating the manipulation problem through a distributed RL paradigm, AcceRL explicitly maximizes long-term rewards rather than mimicking single-step actions. This mechanism allows the policy to recover from minor deviations, ensuring stable execution throughout complex multi-stage tasks and achieving a striking 99.1\% success rate in the Long category.

\subsection{Task Performance on ManiSkill}

To evaluate AcceRL in complex, contact-rich continuous control tasks, we conducted experiments on the ManiSkill PickCube environment (detailed configurations and hyperparameters are provided in Appendix~\ref{app:experiment_setup}, Table~\ref{tab:hyperparameters}), which demands precise physical interaction and dynamic feedback.

As shown in Figure~\ref{fig:maniskill_sr}, the supervised fine-tuning baseline, OpenVLA-OFT, tends to plateau at a success rate of 66\%--80\% under our ManiSkill PickCube setting. In contrast, AcceRL reaches a peak success rate of approximately 90\% within 60k training steps, showing clear gains over supervised fine-tuning. This improvement highlights the benefit of online RL for contact-rich manipulation, where policies must recover from compounding errors and adapt to physical interaction feedback. Rather than merely imitating isolated single-step actions, AcceRL directly optimizes long-horizon task rewards through distributed environment interaction.

\subsection{Performance Evaluation of AcceRL-WM on the LIBERO}

    
    
\FloatBarrier

\begin{figure}[htbp]
    \centering
    
    \begin{subfigure}{0.32\textwidth}
        \centering
        \includegraphics[width=\linewidth]{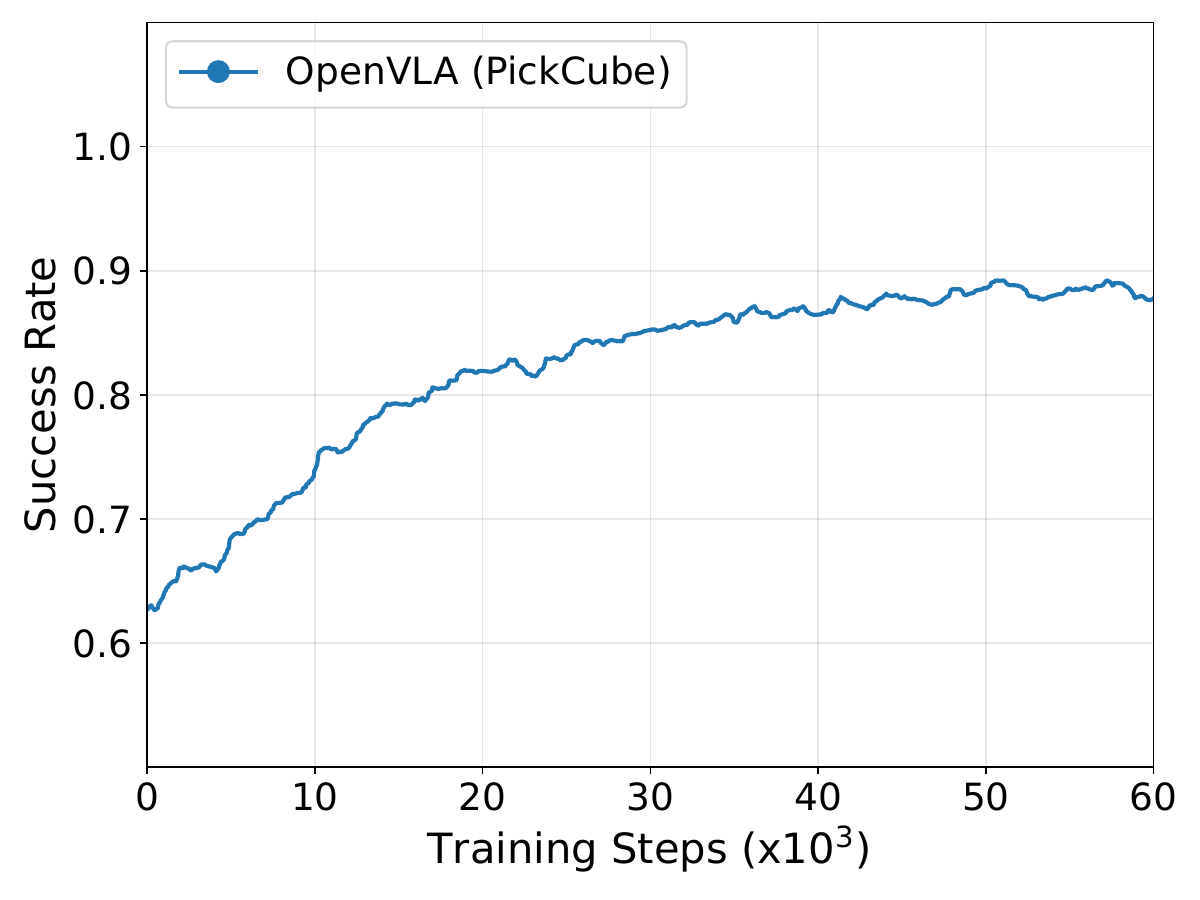}
        \caption{\textbf{AcceRL ManiSkill}}
        \label{fig:maniskill_sr}
    \end{subfigure}\hfill
    \begin{subfigure}{0.32\textwidth}
        \centering
        \includegraphics[width=\linewidth]{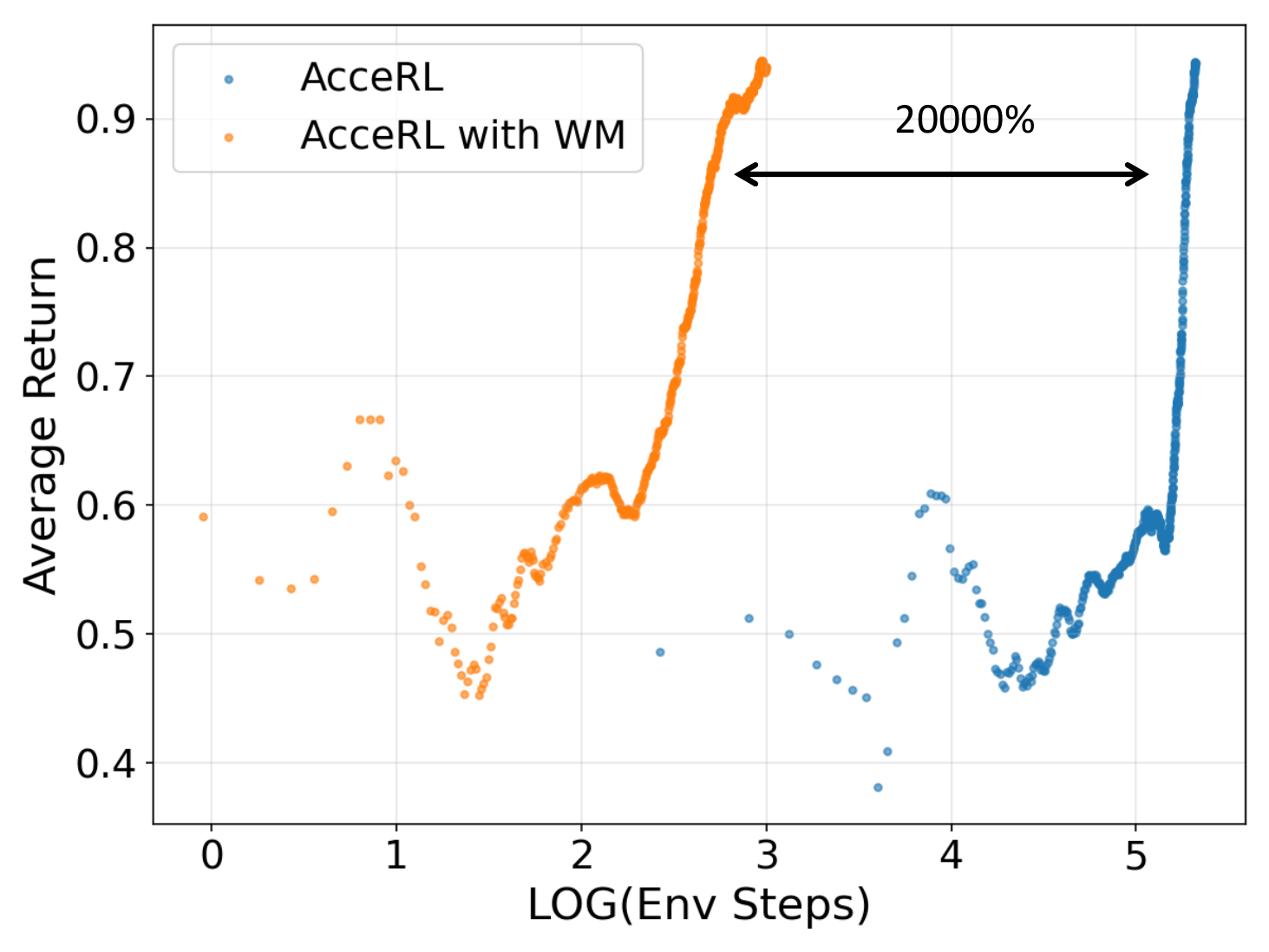}
        \caption{\textbf{AcceRL-WM Diamond}}
        \label{fig:optimized_sr}
    \end{subfigure}\hfill
    \begin{subfigure}{0.32\textwidth}
        \centering
        \includegraphics[width=\linewidth]{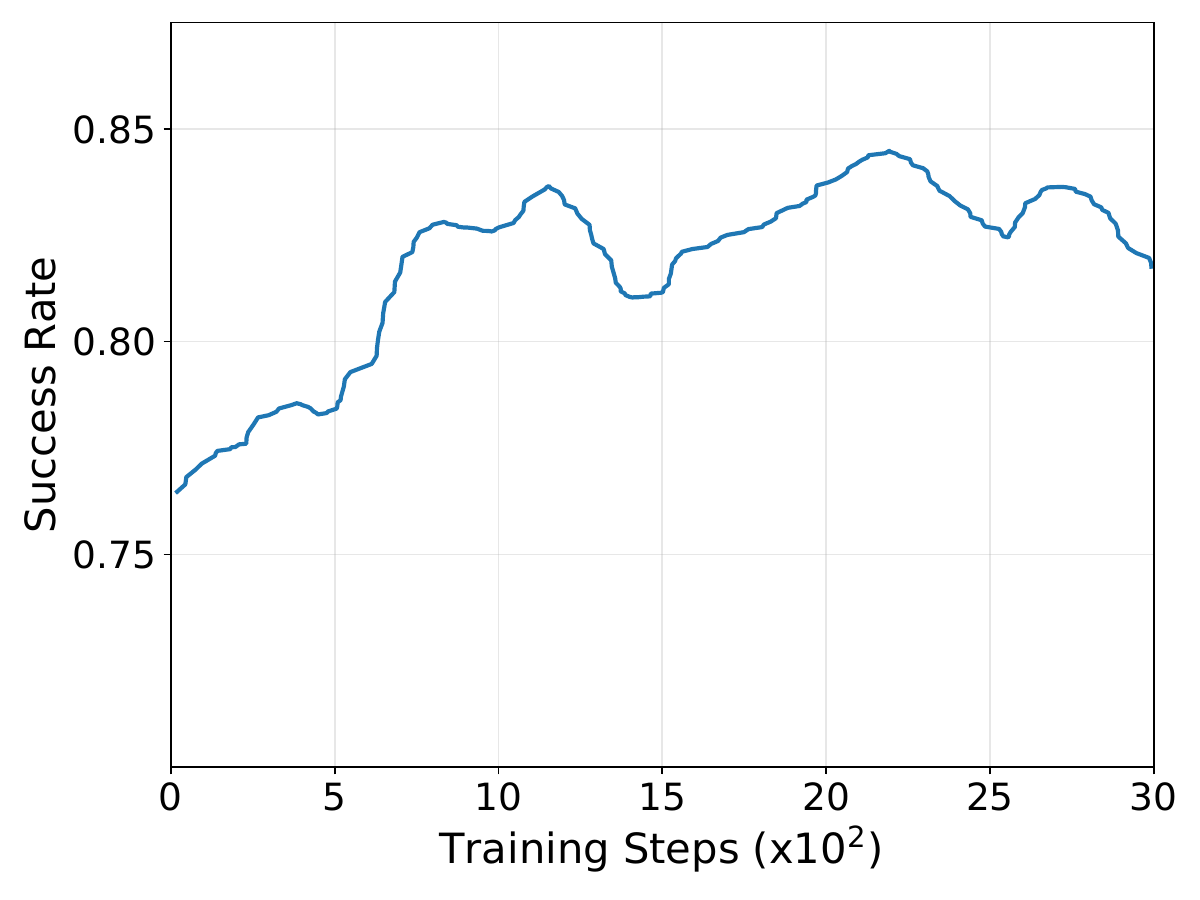}
        \caption{\textbf{AcceRL-WM Cosmos}}
        \label{fig:accerl_cosmos_libero_curve}
    \end{subfigure}
    
    \caption{(a) On ManiSkill PickCube, AcceRL reaches a $\sim$90\% peak success rate, overcoming the early plateaus of supervised baselines. (b) On LIBERO-Spatial, AcceRL-WM demonstrates substantially higher sample efficiency than the model-free AcceRL baseline by reaching a comparable average return with far fewer real environment interactions, corresponding to an approximately 200$\times$ reduction in required environment steps.(c) Replacing the DIAMOND world model with Cosmos, AcceRL-Cosmos successfully completes the closed-loop imagined rollout and policy-update pipeline, suggesting that AcceRL can accommodate alternative world-model backends with limited engineering changes.}
    \label{fig:combined_performance_and_adaptability}
\end{figure}

To evaluate the learning efficiency of our model-based architecture, we initialize the policy with a suboptimal checkpoint (pre-trained on limited demonstrations). As illustrated in Figure~\ref{fig:optimized_sr}, integrating the DIAMOND~\cite{alonso2024diffusion} world model, which is pre-trained on 2,000 out-of-distribution offline trajectories, drastically accelerates VLA fine-tuning. The learning curves demonstrate that AcceRL with the world model requires significantly fewer real environment interactions to achieve a high average return compared to the model-free AcceRL baseline. Specifically, the horizontal gap on the logarithmic x-axis reveals an approximately $200 \times$ (20,000\%) improvement in online sample efficiency. This massive reduction in physical sampling cost is fundamentally driven by the world model's ability to substitute costly real-world interactions with extensive imagination rollouts. Furthermore, leveraging dense potential-based reward signals computed for every imagined transition effectively alleviates the sparse feedback typical of traditional RL and provides an informative optimization landscape.

\subsection{Generalizing AcceRL to the Cosmos World Model}

To further validate the pluggability of AcceRL, we replace the original DIAMOND world model with the Cosmos diffusion world model~\cite{agarwal2025cosmos} while keeping the OpenVLA-OFT policy and the main RL training pipeline unchanged. Given historical multi-view observations and future action sequences, Cosmos predicts subsequent visual states, enabling pixel-space imagined rollouts for policy fine-tuning.

Since Cosmos introduces higher computational cost and different dependency requirements, we deploy its components in isolated Ray~\cite{moritz2018ray} actors and schedule them within the same distributed AcceRL framework. As shown in Figure~\ref{fig:accerl_cosmos_libero_curve}, AcceRL-Cosmos successfully completes imagined rollouts and subsequent policy updates on LIBERO-Spatial, demonstrating that AcceRL can flexibly support heterogeneous world model backends with minor pipeline modification.

%% file: conclusion.tex
\section{Conclusion}
This paper presents AcceRL, an asynchronous reinforcement learning framework optimized for large scale VLA. The core strength of the framework lies in its trainable and plug and play world model module. This integration facilitates fidelity imagination rollouts that allow the system to transcend the frequency constraints of physical simulators while significantly enhancing sample efficiency. Furthermore AcceRL achieves superior performance by leveraging a fully decoupled architecture for rollout and inference and training. This design effectively eliminates systemic bottlenecks such as synchronization overhead and memory constraints. By ensuring high hardware utilization and effectively masking communication overheads, it maximizes the benefits of distributed memory optimizations, unlocking near-linear scalability for rollouts and super-linear throughput for the trainer. By simultaneously improving computational throughput and online sample efficiency, AcceRL addresses key system bottlenecks relevant to physical robotic learning, including data collection cost and wall-clock training time. While our current validation is conducted in simulation, these results point toward a scalable foundation for future real-world robotic RL. Future work will focus on deploying the framework for direct real-world reinforcement learning and extending the architecture to accommodate complex agentic training. 

%% file: appendix.tex
\clearpage
\section{Key Contributions of AcceRL}
\begin{figure}[H]
    \centering
    \includegraphics[width=\textwidth]{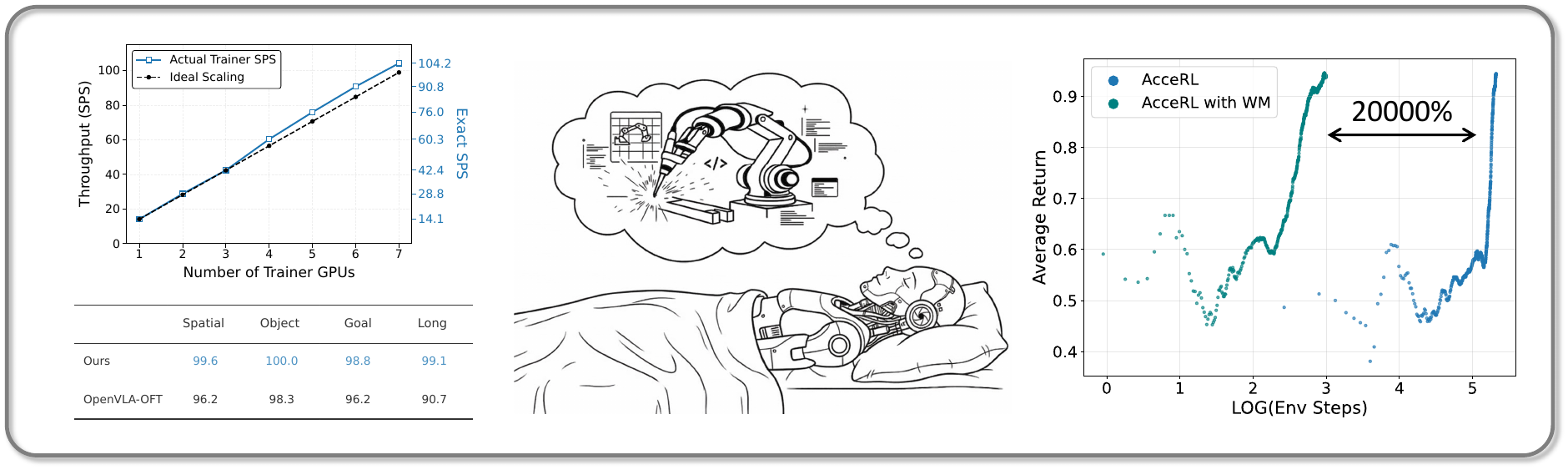}
    \caption{\textbf{Overview of AcceRL}. (Top Left) The framework's throughput exhibits super-linear scaling with the number of trainer GPUs, enabled by ZeRO~\cite{rajbhandari2020zero} optimizations. (Bottom Left) Experimental results on the LIBERO benchmark demonstrate that AcceRL achieves highly competitive performance across all evaluation categories, attaining the best results among the compared methods on Spatial, Object, and Long tasks. (Middle) A conceptual illustration of the world model in reinforcement learning, representing the agent ``learning in imagination'' to significantly boost data utilization. (Right) Performance comparison between model-based (AcceRL with WM) and model-free (AcceRL) approaches. By leveraging a world model pre-trained on 1,000 offline trajectories, the model-based architecture improves online sample efficiency by $200\times$.}
    \label{fig:overview}   
\end{figure}

\section{Timeline Comparison of synchronous RL framework and asynchronous framework AcceRL}
\begin{figure}[H]
    \centering
    \includegraphics[width=\textwidth]{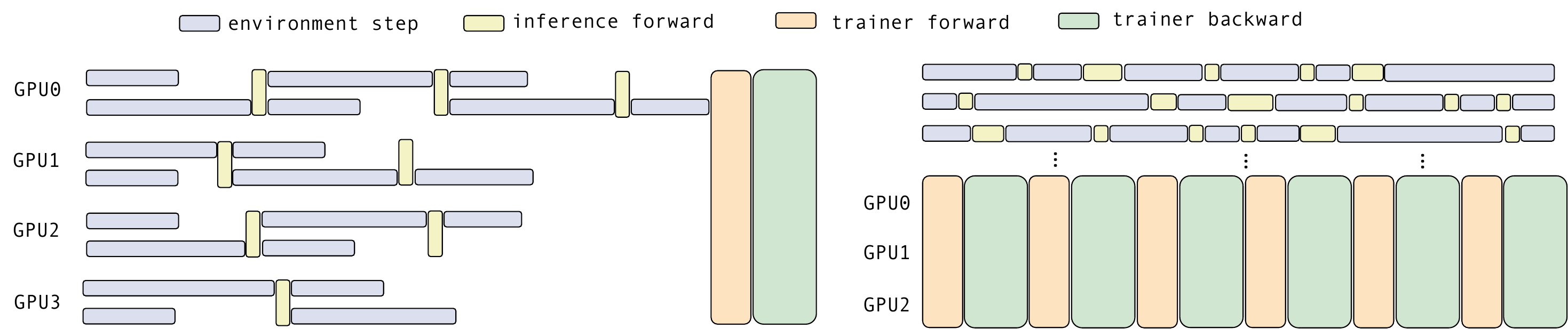}
    \caption{\textbf{Timeline of a synchronous RL framework (left) and the asynchronous framework AcceRL (right)}. The inference forward blocks in AcceRL encompass both the batching wait time and the inference time. By eliminating the GPU bubbles, AcceRL maximizes hardware utilization and significantly enhances overall training efficiency.}
    \label{fig:bubble}   
\end{figure}

\section{More Details about Addressing Policy Lag}
\label{sec:addressing_policy_lag_supply}

In an asynchronous distributed architecture, a significant challenge is policy lag, where the behavior policy $\mu$ used during rollout becomes stale compared to the current learner policy $\pi$. To mitigate the resulting off-policy bias and stabilize convergence, the TrainerWorker implements Value Re-computation, Global Advantage Normalization, and Gradient Calibration.

\subsection{Theoretical Equivalency and Implementation Details of Value Recomputation}
\label{sec:value_recomputation_suppl}

\paragraph{Deterministic Micro-Batch Slicing}
Regarding the rigor of deterministic micro-batch slicing, considering that the system employs a gradient accumulation mechanism, the model weights remain frozen until the processing of the entire large batch is complete. According to the linear superposition property of gradients, the gradient of the total objective function can be expressed as the cumulative sum of the gradients of each micro-batch:
\begin{equation}
    \nabla_{\theta} \mathcal{J}_{total} = \sum_{i=1}^{N_{acc}} \nabla_{\theta} \mathcal{J}_{mini\_batch}^{(i)}
\end{equation}
where $N_{acc}$ represents the number of accumulation steps.Because model parameters remain fixed within each gradient-accumulation window, reusing the training forward pass for GAE computation does not introduce additional parameter staleness. Given the same set of trajectory samples before each optimizer step, sequential micro-batch slicing preserves the same large-batch objective under exact arithmetic while avoiding redundant shuffling and value-forward computation. 

\paragraph{Communication-Hiding Lag Normalization}
Regarding lag normalization for communication hiding, the system uses statistics from the previous step to standardize the advantage function:
\begin{equation}
    \hat{A}_t = \frac{A_t - \mu_{A, t-1}}{\sigma_{A, t-1} + \epsilon}
\end{equation}
where $\mu_{A, t-1}$ and $\sigma_{A, t-1}$ denote the global mean and standard deviation of advantages from the previous optimization step.
This design provides a practical approximation for hiding the synchronization latency of global advantage normalization. Rather than blocking training to wait for the current batch statistics, AcceRL uses the global statistics from the previous optimization step. The approximation assumes that adjacent-batch advantage statistics change smoothly in practice, which is encouraged by bounded policy updates and large distributed batches but is not theoretically guaranteed. Empirically, we observe that this delayed normalization introduces limited discrepancy while preserving stable convergence.


\paragraph{Engineering Implementation}
At the engineering implementation level, low-overhead just-in-time Generalized Advantage Estimation (GAE) requires precise handling of boundary states after the micro-batch forward pass. If the sampled data is an unterminated episode, to compute the GAE, the state value of the last time step serves solely as a bootstrap target. Its corresponding loss function is forcibly set to zero to block gradient backpropagation, and the target value node must be detached from the computation graph. During the asynchronous global synchronization phase, the system silently records the sum and the sum of squares of the statistics for the current batch within the local computation loop. Once the synchronization operations across compute nodes are completed at the gradient accumulation boundary, the system uniformly updates the global sliding state using Welford's algorithm. This ensures the numerical stability of streaming variance calculations and makes them available for the next time step.


\subsection{Global Advantage Normalization}
To ensure training stability across the distributed training architecture, AcceRL implements a global normalization process after advantage recomputation. Since sampling noise and task difficulties vary across different nodes, utilizing global statistics is essential to eliminate local biases and provide a unified baseline for policy updates.

Instead of performing costly global data shuffling, we efficiently aggregate local statistics (sum, squared sum, and element count) from all nodes via a single distributed \texttt{AllReduce} operation. The pseudo-code for our communication-efficient normalization is as follows:

\begin{lstlisting}[language=Python, basicstyle=\ttfamily\small,
                   keywordstyle=\color{blue},
                   commentstyle=\color{gray},
                   frame=single]
import torch
import torch.distributed as dist

def global_advantage_norm(adv, eps=1e-8):
    # 1. Compute local statistics
    local_sum = adv.sum()
    local_sqr_sum = (adv ** 2).sum()
    local_count = torch.tensor(
        adv.numel(), device=adv.device, dtype=adv.dtype
    )

    # 2. Pack tensors for a single communication call
    stats = torch.stack([local_sum, local_sqr_sum, local_count])

    # 3. Synchronous AllReduce across all nodes
    dist.all_reduce(stats, op=dist.ReduceOp.SUM)

    # 4. Unpack and compute global mean and standard deviation
    global_sum, global_sqr_sum, global_count = stats
    global_mean = global_sum / global_count
    global_var = (global_sqr_sum / global_count) - (global_mean ** 2)
    global_std = torch.sqrt(torch.clamp(global_var, min=0.0))

    # 5. Standardize advantages
    return (adv - global_mean) / (global_std + eps)
\end{lstlisting}

The standardized advantage $\hat{A}_t$ is subsequently used for policy gradient computation, ensuring stable optimization signals and improving convergence robustness in high-dimensional manipulation tasks.

\input{refinements} 

\section{Experiment Setup}
\label{app:experiment_setup}
The framework consists of three core components for world-model-based reinforcement learning. The policy model $M_{policy}$ utilizes the OpenVLA-OFT~\cite{kim2025fine} architecture to process observation images and proprioceptive states into discretized action tokens and employs action chunking for temporal consistency. The observation model $M_{obs}$ is a diffusion-based transition model built upon the DIAMOND (DIffusion As a Model Of eNvironment Dreams) ~\cite{alonso2024diffusion} architecture. Unlike traditional world models that compress environment dynamics into discrete latent variables, DIAMOND leverages a diffusion process to preserve fine-grained visual details. This enables $M_{obs}$ to generate high-fidelity, synthesized next-frame images conditioned on historical observation sequences and current action chunks. Acting as a virtual referee, the reward model $M_{reward}$ is a binary classifier modified from OpenVLA-OFT that evaluates these synthetic frames to estimate success probabilities and termination signals.In our experiments, we utilize the OpenVLA-OFT backbone and the DIAMOND world model. The OpenVLA codebase is released under the MIT License, while its pre-trained model weights (built upon Llama-2-7B) are distributed under the Llama 2 Community License. DIAMOND and other baseline implementations are used in accordance with the MIT License.

Our experiments were conducted on high-performance compute nodes equipped with NVIDIA H200 (141GB) GPUs, providing the necessary VRAM for efficient VLA fine-tuning. The ManiSkill training runs utilized 4 GPUs for 2.193 days (totaling $\sim$210.5 GPU-hours), while the LIBERO benchmark training utilized 8 GPUs for 2 days (totaling $\sim$384 GPU-hours). The total compute consumption for the core results is approximately 594.5 H200 GPU-hours.

\clearpage

\makeatletter
\setlength{\@fptop}{0pt} 
\makeatother

\begin{table}[p]
\centering
\caption{Detailed Experimental Settings and Hyperparameters for AcceRL on ManiSkill tasks.}
\label{tab:hyperparameters}
\begin{tabular}{@{}ll@{}}
\toprule
\textbf{Category} & \textbf{Value} \\ \midrule
\textbf{Environment \& Task} & \\
Simulator & ManiSkill (Task: PickCube-v1) \\
Observation Space & RGB Camera (\texttt{base\_camera}, 128$\times$128) \\
Max Episode Steps & 100 \\ \midrule
\textbf{Model Architecture} & \\
VLA Backbone & OpenVLA-OFT (based on Llama-2-7B) \\
Action Vocabulary Size & 256 bins (Slimmed) \\
Value Head & MLP with Action-Aware Attention Pooling \\ \midrule
\textbf{RL Training (GIPO)} & \\
Optimizer & AdamW (with bfloat16 mixed precision) \\
Learning Rate (Policy / Value) & $3 \times 10^{-6}$ / $3 \times 10^{-5}$ \\
Warmup Steps & 500 \\
Micro-batch Size & 16 per GPU \\
Gradient Accumulation Steps & 2 \\
Total Global Batch Size & 96 \\
Discount Factor ($\gamma$) & 0.99 \\
GAE Lambda ($\lambda$) & 0.95 \\
GIPO Clip Epsilon & 0.2 \\
KL Divergence Coefficient & 0.1 \\
Entropy Coefficient & 0.0 \\ \midrule
\textbf{AcceRL System Settings} & \\
Value Recomputation & Enabled (\texttt{recompute\_value: true}) \\
Trainer GPUs & 3 \\
Inference GPUs & 1 \\
CPU Rollout Workers & 6 \\
Replay Buffer Capacity & 3000 episodes \\
Total Training Iterations & 60,000 \\
\bottomrule
\end{tabular}
\end{table}
\clearpage

\clearpage

\makeatletter
\setlength{\@fptop}{0pt} 
\makeatother

\begin{table}[p]
\centering
\caption{Detailed Experimental Settings and Hyperparameters for OpenVLA-OFT RL with Cosmos WM on LIBERO Object tasks.}
\label{tab:openvla_cosmos_wm_hyperparameters}
\begin{tabular}{@{}ll@{}}
\toprule
\textbf{Category} & \textbf{Value} \\ \midrule
\textbf{Environment \& Task} & \\
Simulator / Benchmark & LIBERO \\
Observation Space & RGB Image (\texttt{num\_images\_in\_input:2}) \\
\midrule
\textbf{Model Architecture} & \\
VLA Backbone & OpenVLA-OFT \\

Cosmos WM Actions & 16 \\ \midrule
\textbf{RL Training} & \\
Mixed Precision & bfloat16 (\texttt{--use-bf16}) \\
Learning Rate (Policy / Value / Reward / Denoiser) & $1 \times 10^{-5}$ / $1 \times 10^{-4}$ / $1 \times 10^{-4}$ / $1 \times 10^{-5}$ \\
Warmup Steps (Policy / Value / Reward / Denoiser) & 500 / 500 / 500 / 500 \\
Train Batch Size & 8 \\
Gradient Accumulation Steps & 16 \\
Effective Trainer Batch Size & 128 ($1 \times 8 \times 16$) \\
Discount Factor ($\gamma$) & 0.99 \\
GAE Lambda ($\lambda$) & 0.95 \\
GIPO sigma & 0.2 \\

Value Loss Coefficient & 0.5 \\
KL Divergence Coefficient & 0.1 \\
Maximum KL & 0.02 \\
Entropy Coefficient & 0.01 \\
\midrule
\textbf{World Model \& Cosmos Settings} & \\
Imagine Horizon & 8 \\
Number of Step Conditions & 13 \\
Reward Scale & 5.0 \\
Cosmos State $t$ & 8 \\
Cosmos Guidance Scale & 7.0 \\
Cosmos Sampling Steps & 35 \\
Cosmos Number of Returns & 8 \\
Denoiser Batch Size & 2 \\
Denoiser Accumulation Steps & 2 \\
Denoiser Train Interval & 3 \\
Reward Batch Size & 2 \\
Reward Accumulation Steps & 2 \\
Reward Train Interval & 10 \\ \midrule
\textbf{System Settings} & \\
Trainer GPUs & 1 \\
Inference Actors & 1 \\
Reward Inference Actors & 1 \\
Denoiser Inference Actors & 1 \\
Rollout Workers & 30 \\
Evaluation Workers & 10 \\
Inference Batch Size & 8 \\
Total Training Iterations & 30,000 \\
Replay Capacity & 10,000 \\
World Model Replay Capacity & 50,000 \\
Real Trajectory Collect Interval & 3 \\
\bottomrule
\end{tabular}
\end{table}
\clearpage

\clearpage
\begin{table}[p]
\centering
\caption{Detailed Experimental Settings and Hyperparameters for AcceRL WM on LIBERO tasks.}
\label{tab:accerl_wm_hyperparameters}
\begin{tabular}{@{}ll@{}}
\toprule
\textbf{Category} & \textbf{Value} \\ \midrule
\textbf{Environment \& Task} & \\
Simulator & LIBERO \\
Observation Space & RGB Image (\texttt{num\_images\_in\_input:1}) \\ \midrule
\textbf{Model Architecture} & \\
VLA Backbone & OpenVLA-OFT (based on Llama-2-7B) \\
Action Vocabulary Size & 256 bins (Discrete Actions) \\ \midrule
\textbf{RL Training (GIPO)} & \\
Optimizer & AdamW (with bfloat16 mixed precision) \\
Learning Rate (Policy / Value / Reward /denoiser) & $1 \times 10^{-5}$ / $1 \times 10^{-4}$ / $5 \times 10^{-5}$/ $1 \times 10^{-4}$ \\
Warmup Steps (Policy / Value) & 500 / 500 \\
Micro-batch Size & 16 per GPU \\
Gradient Accumulation Steps & 40 \\
Total Global Batch Size & 1280 ($2 \times 16 \times 40$) \\
Discount Factor ($\gamma$) & 0.99 \\
GAE Lambda ($\lambda$) & 0.95 \\
GIPO sigma & 0.2 \\
KL Divergence Coefficient & 0.1 \\
Entropy Coefficient & 0.00 \\ \midrule
\textbf{World Model \& System Settings} & \\
Imagine Horizon & 2 \\
Number of Step Conditions & 4 \\
Reward Train Interval & 15 \\
Trainer GPUs & 2 \\
Inference Workers & 1 \\
CPU Rollout Workers & 24 \\
Evaluation Workers & 12 \\
Total Training Iterations & 30,000 \\
\bottomrule
\end{tabular}
\end{table}
\clearpage

\clearpage
\begin{table}[p]
\centering
\caption{Detailed Experimental Settings and Hyperparameters for OpenVLA RL Training.}
\label{tab:ppo_hyperparameters}
\begin{tabular}{@{}ll@{}}
\toprule
\textbf{Category} & \textbf{Value} \\ \midrule
\textbf{Environment \& Task} & \\
Benchmark & LIBERO \\
Observation Space & RGB images + proprioceptive state \\
Number of Images in Input & 2 \\
\midrule

\textbf{Model Architecture} & \\
VLA Backbone & OpenVLA-OFT \\
Action Space & Discrete actions \\
\\ \midrule

\textbf{RL Training (GIPO / PPO)} & \\
Optimizer & AdamW (with bfloat16 mixed precision) \\
Learning Rate (Policy / Value) & $3 \times 10^{-6}$ / $3 \times 10^{-5}$ \\
Train Batch Size & 8 \\
Gradient Accumulation Steps & 16 \\
Total Training Iterations & 60,000 \\
Discount Factor ($\gamma$) & 0.99 \\
GAE Lambda ($\lambda$) & 0.95 \\
Value Function Coefficient & 0.5 \\
KL Divergence Coefficient & 0.1 \\
Entropy Coefficient & 0.0 \\
GIPO Sigma & 0.5 \\ \midrule

\textbf{System Settings} & \\
Trainer GPUs & 3 \\
Inference Actors & 1 \\
CPU Rollout Workers & 30 \\
Evaluation Workers & 10 \\
Inference Batch Size & 8 \\
Replay Buffer Capacity & 3000 episodes \\
Object Store Memory & 256 GB \\
Checkpoint Interval & 2,000,000 steps \\
\bottomrule
\end{tabular}
\end{table}
\clearpage
\section{Detailed Tabular Data Comparison of Throughput}
\begin{table}[htbp]
    \centering
    \caption{\textbf{Scaling performance and hardware utilization of AcceRL.} The data illustrates a super-linear increase in training throughput (SPS) while maintaining consistently high GPU utilization ($>94\%$) at different scale points.}
    \label{tab:gpu_scaling_util}
    \begin{tabular}{ccc}
        \toprule
        \textbf{GPUs}  & \textbf{Throughput (SPS)} & \textbf{GPU Util. (\%)} \\
        \midrule
        1  & 14.13 & 96.45 \\
        2  & 28.82 & 97.17 \\
        3  & 42.42 & 94.22 \\
        4  & 60.33 & 98.36 \\
        5  & 75.95 & 96.72 \\
        6  & 90.78 & 96.63 \\
        7  & 104.22 & 95.07 \\
        \bottomrule
    \end{tabular}
\end{table}
\FloatBarrier

\section{Ablation Study}
\subsection{Ablation Study of AcceRL without Value Recomputation}
\label{sec:revalue_ablation}

\begin{figure}[htbp]
    \centering
    \includegraphics[width=0.5\textwidth]{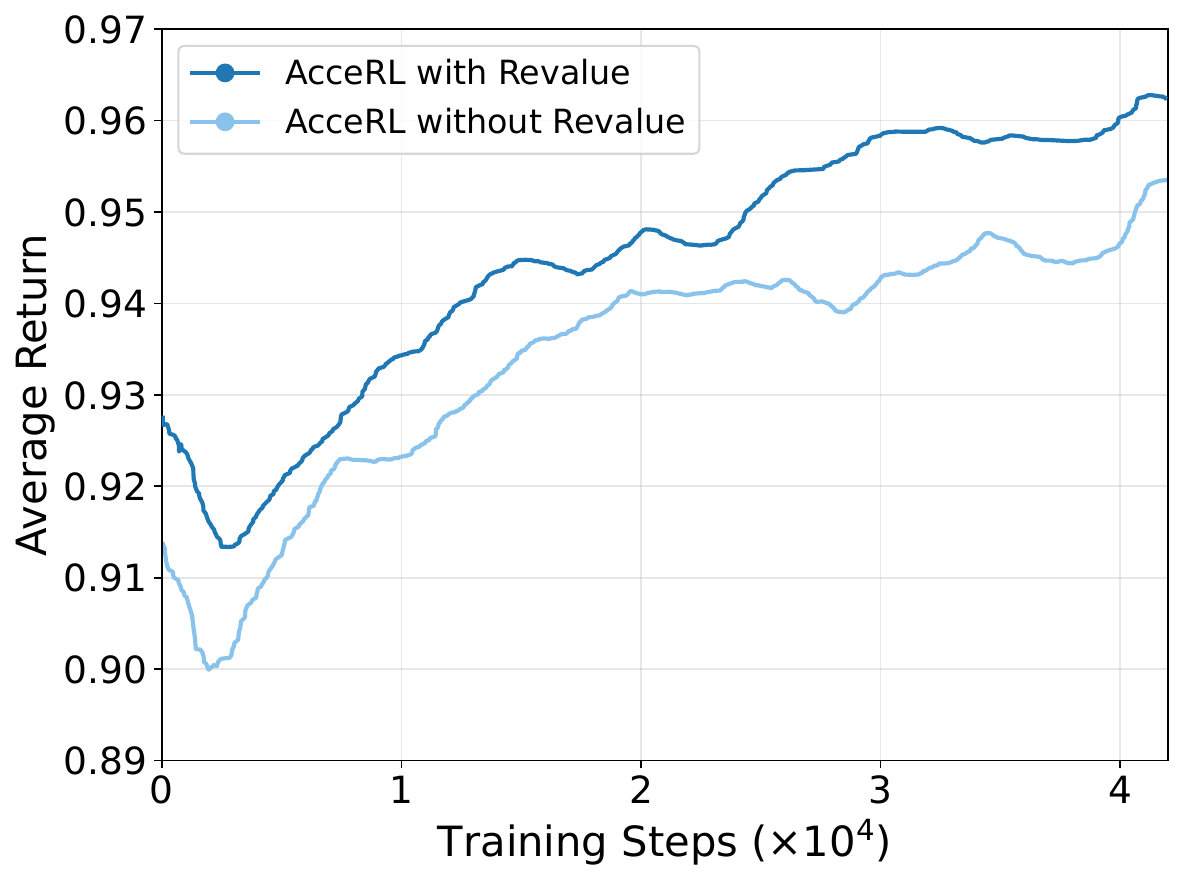}
    \caption{\textbf{Ablation study of the AcceRL framework regarding the value recomputation mechanism.} The comparison highlights the critical role of recomputing values in maintaining training stability. AcceRL with Revalue (dark blue) effectively handles data staleness in low-rollout scenarios, while AcceRL without Revalue (light blue) exhibits significant performance drops and higher variance due to misaligned value targets.}
    \label{fig:revalue_ablation}
\end{figure}
\FloatBarrier

While the previous results demonstrate the overall efficiency of the AcceRL framework, its performance in asynchronous distributed settings relies heavily on maintaining consistency between the actor and the critic. A significant technical challenge in such architectures is data staleness: since rollout and training occur in parallel, the experiences stored in the replay buffer were often evaluated by an earlier, outdated version of the critic network. To mitigate the resulting gradient noise and training instability, we introduced a value recomputation mechanism that updates value targets using the most recent critic before each gradient update. 
As illustrated in Figure~\ref{fig:revalue_ablation}, the value recomputation mechanism serves as a critical stabilizer for off-policy VLA fine-tuning. The full AcceRL framework (dark blue) effectively handles data staleness, maintaining high average returns with minimal variance even in resource-constrained settings. 

In contrast, the version without recomputation (light blue) exhibits performance oscillations and a lower convergence ceiling. This instability arises from misaligned value targets, where the policy is updated based on evaluations that no longer reflect the current state of the critic. By ensuring that the policy is always guided by the most accurate and up-to-date value estimates, the recomputation mechanism prevents error accumulation and ensures robust policy improvement throughput the training process.
\subsection{Ablation Study of Gaussian Importance sampling Policy Optimization}
\label{sec:gipo_ablation}

Beyond system-level optimizations, the choice of policy optimization objective is critical for stability in asynchronous settings.Given that the inherent policy lag in decoupled architectures can destabilize standard PPO, we employ GIPO ~\cite{lu2026gipo}, which replaces hard clipping with a Gaussian trust weight to softly damp extreme importance ratios.

\begin{figure}[htbp]
    \centering
    \includegraphics[width=0.5\textwidth]{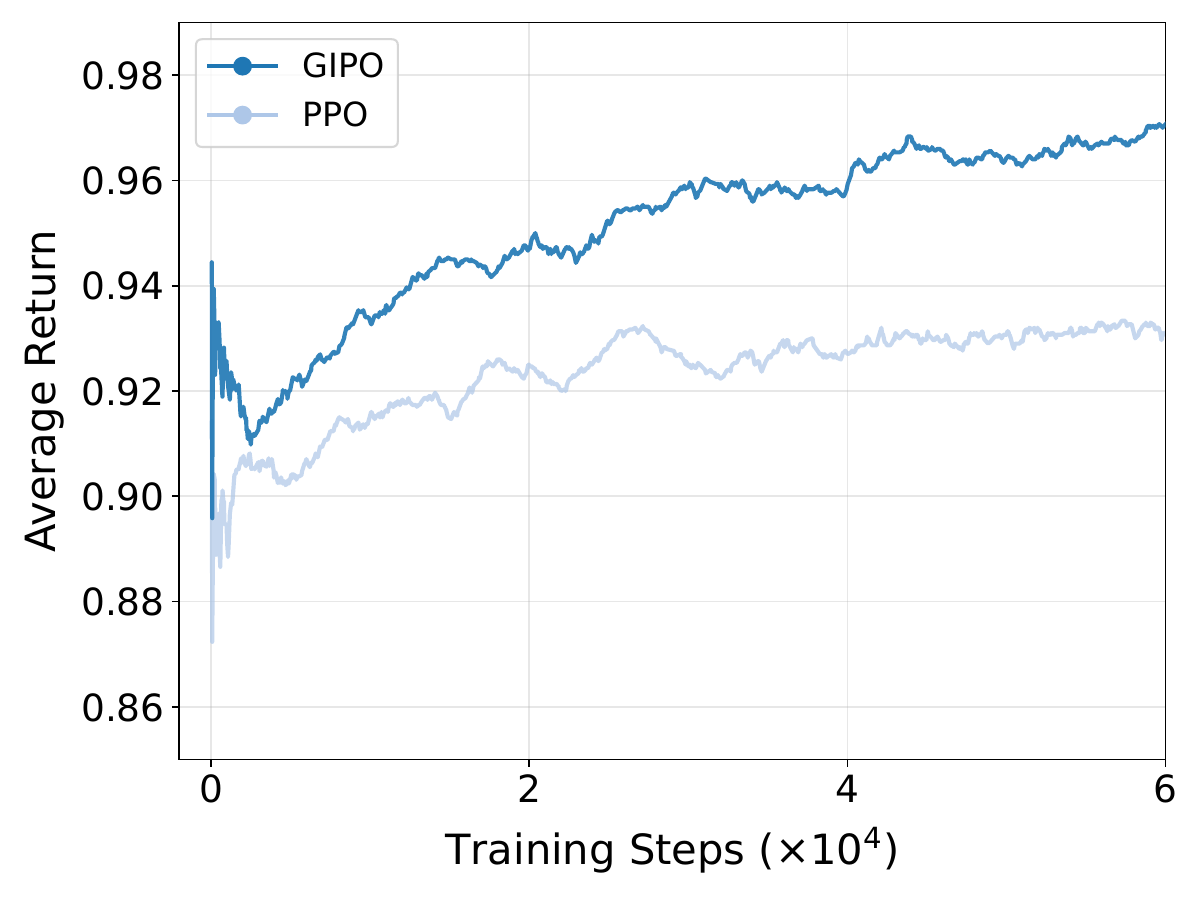}
    \caption{\textbf{Ablation study of the AcceRL framework regarding the GIPO algorithm.} The curves highlight the stability of GIPO (dark blue) compared to standard PPO (light blue). GIPO maintains a robust, high-performance trajectory, whereas PPO exhibited oscillatory behavior and inferior performance due to policy lag.}
    \label{fig:gipo_ablation}
\end{figure}

As illustrated in Figure~\ref{fig:gipo_ablation}, we plot the raw evaluation returns to explicitly visualize the training dynamics. A prominent, rapid surge is observed at the very beginning of the training (near 0k steps). This phenomenon is a statistical artifact inherent to the Episode-Level Long-Tail in our asynchronous rollout mechanism. During the initial phase, successful trajectories typically have shorter horizons and are therefore completed and reported to the buffer faster than unsuccessful ones that persist through the full task duration ~\cite{cipar2013solving}. This leads to a temporary, optimistic skew in success rates before the system reaches a representative sample distribution.

\FloatBarrier

Once stabilized, the comparison reveals a clear divergence. The AcceRL framework equipped with GIPO (dark blue) maintains a tightly clustered, high-performance trajectory. In contrast, the AcceRL variant utilizing standard PPO (light blue) exhibits severe performance oscillations and significantly lower sample efficiency. This instability is primarily due to PPO's reliance on a hard clipping mechanism, which becomes counterproductive in asynchronous settings where data generation can be delayed or experiences may become stale. In such scenarios, a large proportion of the sampled data triggers the clipping threshold, resulting in ineffective stale updates with zero gradients that fail to improve the policy. 

To overcome these algorithmic limitations, GIPO replaces hard clipping with a smooth log-ratio trust-weighted surrogate. This design inherently provides a symmetric and continuous penalty, implicitly enforcing a tunable bound on the policy update magnitude without abruptly zeroing out useful gradients. Furthermore, by achieving a superior bias-variance trade-off, GIPO provides formal guarantees of robustness and stability under finite-sample estimation, even when learning from highly stale replay data. By effectively mitigating this data utilization collapse, GIPO demonstrates a dramatic acceleration in convergence: it achieves a performance level at approximately 8,000 steps that standard PPO requires 60,000 steps to reach, representing a roughly 7.5-fold improvement in sample efficiency.

\subsection{Ablation Study of the Trainer-Infer Weight Synchronization Mechanism}
To analyze the impact of synchronization efficiency on system performance, we compare our framework's NCCL collective communication scheme with two other common synchronization paths in distributed reinforcement learning:
\begin{enumerate}
    \item \textbf{Host-Mediated PCIe Transfer}: This approach is also known as CPU-staged GPU memory transfer. In a typical workflow, the trainer copies weights from GPU memory to CPU host memory. These weights are then distributed to inference nodes via a parameter server, Ray Object Store, or underlying TCP paths. Finally, the Inference Workers copy the weights from host memory back to their GPU memory. In this path, PCIe bandwidth, object management overhead, and potential Python serialization become the primary bottlenecks.
    \item \textbf{Shared-Storage Checkpoint Reload}: Methods like AReal couple policy synchronization with checkpoint management. After completing a training step, the trainer saves the model weights to a shared disk (such as NFS or a parallel file system). The inference servers then use polling signals to trigger a disk read and reload the model.
\end{enumerate}
Our experiments are conducted on four 8-GPU H200 nodes, with the 7B OpenVLA-OFT as the backbone model. We allocate 6 GPUs for training and 2 GPUs for inference.

\begin{table}[htbp]
    \centering
    \caption{\textbf{Comparison of synchronization overheads across different methods.} By employing NCCL with an early drain strategy, our system significantly reduces synchronization latency, infer update delay, and sample policy lag compared to traditional host-mediated or shared-storage approaches.}
    \label{tab:sync_methods}
    \begin{tabular}{lccc}
        \toprule
        \textbf{Method} & \textbf{Sync Latency} & \textbf{Infer Update Delay} & \textbf{Sample Policy Lag} \\
        \midrule
        NCCL with drain & \textbf{70--95 ms} & \textbf{70--95 ms} & \textbf{0.03--0.05} \\
        \addlinespace
        NCCL without drain & 220--300 ms & 220--300 ms & 0.1--0.2 \\
        \addlinespace
        PCIe / Host-mediated & 2--7 s & 3--12 s & 0.3--1.5 \\
        \addlinespace
        Shared-storage / AReal-style & 12--53 s & 17--76 s & 2--12 \\
        \bottomrule
    \end{tabular}
\end{table}

Experimental results demonstrate the advantages of the NCCL-based synchronization mechanism for online training of ultra-large-scale VLA models. By eliminating redundant disk I/O and CPU transfers, this approach successfully reduces the update latency for 7B-scale models from tens of seconds to a sub-second level of under 100 milliseconds, achieving a performance improvement of two orders of magnitude.
Combined with inference drain, the system further optimizes the trainer's waiting overhead at synchronization points, significantly reducing the sync wait time from 220–300 ms to 70–95 ms. This fine-grained pipeline alignment ensures high throughput on the training side.
Another impact is reflected in the significant reduction of policy staleness. Experiments show that a more efficient communication path directly reduces the Sample Policy Lag, which drops from 2–12 under the shared storage scheme to 0.03–0.05. This means rollout data generated by Inference Workers can quickly reflect the policy distribution currently optimized by the trainer. This fast feedback is critical for large-scale VLA models to maintain numerical stability and convergence efficiency in complex physical simulation tasks.
Furthermore, compared to AReal, which suffer from global barrier blocking and severe long-tail I/O latency during synchronous writes, our NCCL synchronization scheme aligns perfectly with the distributed training pipeline. Even in high-frequency synchronization scenarios, the proposed mechanism exhibits excellent system scalability, with nearly negligible impact on overall training throughput.
\subsection{Comprehensive Multi-Seed Evaluation}
\label{sec:comprehensive_eval}

To rigorously evaluate the statistical significance and competitive upper bound of our proposed mechanisms, we conducted an extensive multi-seed evaluation across 10 MuJoCo environments. Benchmarking five algorithmic variants—standard PPO, SAPO, and GIPO with varying base trust region widths ($\sigma \in \{0.2, 0.5, 1.0\}$)—we collected valid trajectories from 196 successful independent runs (accounting for minor cluster preemptions out of 200 theoretical runs). Following the rigorous protocols in \textit{RLiable}, we report the Interquartile Mean (IQM) and Mean Normalized Scores based on the final 50 evaluation points of the training process. The IQM provides a robust aggregate metric by discarding the highest and lowest 25\% of runs, effectively filtering out anomalous seed performances to reveal the algorithm's stable learning capability.

\begin{table}[htbp]
    \centering
    \caption{\textbf{Aggregate Performance Scores.} Evaluated across 196 independent runs in 10 MuJoCo environments. GIPO(0.2, 1.0) achieves a substantial lead in both IQM (robustness) and Mean Normalized score (overall average).}
    \label{tab:aggregate_iqm_norm}
    \begin{tabular}{llcc}
        \toprule
        \textbf{Rank} & \textbf{Algorithm} & \textbf{IQM} & \textbf{Mean Norm.} \\
        \midrule
        1 & \textbf{GIPO(0.2, 1.0)} & \textbf{0.724} & \textbf{0.651} \\
        2 & PPO & 0.595 & 0.565 \\
        3 & GIPO(0.5, 1.0) & 0.578 & 0.582 \\
        4 & GIPO(1.0, 1.0) & 0.446 & 0.481 \\
        5 & SAPO & 0.315 & 0.366 \\
        \bottomrule
    \end{tabular}
\end{table}
\FloatBarrier

As detailed in Table~\ref{tab:aggregate_iqm_norm}, GIPO(0.2, 1.0) establishes a commanding lead across all metrics, achieving an IQM of 0.724 and a Mean Normalized score of 0.651. This creates a substantial performance gap over the baseline PPO (0.595 IQM) and completely outperforms SAPO (0.315 IQM). 

These comprehensive results provide critical insight into asynchronous training dynamics: under conditions of high data staleness, a narrower foundational trust region ($\sigma=0.2$) serves as an essential stabilizer. While standard PPO struggles with policy lag due to hard clipping, and larger $\sigma$ values (e.g., 1.0) allow excessive divergence, GIPO's tightly bounded Gaussian trust weight filters out heavily delayed off-policy data without abruptly zeroing out all gradients. This strict yet smooth attenuation prevents catastrophic policy degradation, enabling GIPO to consistently learn high-performing policies even in challenging, delayed-feedback distributed architectures.

\section{Broader Impacts}

Our work on the AcceRL framework introduces significant efficiency improvements for training Vision-Language-Action (VLA) models. A primary positive impact is the democratization of embodied AI research; by significantly reducing the computational resources and time required to train large-scale models, we lower the barrier to entry for researchers with limited compute budgets, which also helps reduce the carbon footprint associated with heavy GPU training. Regarding potential negative impacts, accelerating the development of highly capable robotic policies could inadvertently expedite labor automation, potentially leading to workforce displacement in certain sectors. Furthermore, as with any embodied AI system, deploying these models in real-world physical environments without rigorous safety guardrails poses inherent physical risks.


%% file: refinements.tex
\section{Implementation Details and Refinements for VLA RL}
\label{sec:refinements}
\subsection{Vocabulary Slimming}
To efficiently adapt the OpenVLA-OFT ~\cite{kim2025fine} (based on the Llama-2-7B language backbone) for RL, we prune the language model head $lm\_head$ to output only the action vocabulary (e.g., 256 bins), decoupling the policy from the full language lexicon. This reduces VRAM and compute costs while preventing probability mass from leaking to non-action tokens.

AcceRL employs an in-place replacement strategy to crop the original, massive output layer $lm\_head$ into a compact linear layer containing only action-related tokens. Specifically, the system first locates the start index and end index of the action tokens in the original vocabulary, then clones the corresponding weights from the original $lm\_head.weight$ via slicing. Finally, a new linear layer with an output dimension of $N_{actions}$ is created, and the extracted weights are migrated to this layer, directly overwriting the original language model head.

By slimming the vocabluary, the model only needs to output action commands and does not need to process language text, provides the following advantages:
\begin{itemize}
    \item VRAM Efficiency: Reducing the weight matrix from $[N_{vocab}, d_{model}]$ (where $N_{vocab} \approx 32k$) to $[N_{actions}, d_{model}]$ significantly reduces the existing occupancy of gradients and optimizer states.
    \item Increased throughput: Eliminating most of the computational overhead of the last layer significantly speeds up the iteration frequency of each training step.
    \item Noise Reduction: Physically isolating non-action tokens prevents the model from assigning probability to irrelevant vocabulary, focusing the policy distribution and speeding up convergence. 
\end{itemize}



\subsection{Value Head Design}

The Value Head is designed to estimate the state value $V(o_t)$ by focusing on decision-critical information within the VLA's hidden states. In AcceRL, we implement an Action-Aware Attention Pooling mechanism. Given a sequence of all action-related hidden states extracted from the last hidden layer of the VLA, we compute an attention score for each token. These hidden states are detached from the VLA backbone's computation graph to prevent value gradients from interfering with the policy representation. 

The normalized attention weights are then derived via a softmax operation, and the pooled representation is calculated as a weighted sum. Crucially, a state's value is intrinsically tied to the remaining time horizon; for instance, as the current step approaches the maximum episode limit, the expected future returns drop to zero due to imminent truncation, regardless of the agent's proximity to success. To explicitly account for this temporal dependency, we incorporate a step embedding that maps the current episode step $t$ to a high-dimensional vector, which is element-wise added to the pooled feature. The final state value is estimated through a Multi-Layer Perceptron (MLP). The exact tensor operations and dimension transformations are detailed in the PyTorch implementation below:

\begin{lstlisting}[language=Python, basicstyle=\ttfamily\small, keywordstyle=\color{blue}, commentstyle=\color{gray}, frame=single, breaklines=true]
import torch
import torch.nn as nn
import torch.nn.functional as F

class ActionAwareValueHead(nn.Module):
    def __init__(self, hidden_dim, max_episode_steps):
        super().__init__()
        # Attention projection
        self.attn_proj = nn.Linear(hidden_dim, 1)
        
        # Temporal step embedding
        self.step_emb = nn.Embedding(max_episode_steps, hidden_dim)
        
        # Value estimation MLP
        self.mlp = nn.Sequential(
            nn.Linear(hidden_dim, hidden_dim),
            nn.GELU(),
            nn.Linear(hidden_dim, 1)
        )

    def forward(self, hidden_states, step_t):
        # hidden_states: [Batch, Seq_Len, Hidden_Dim]
        # step_t: Current episode step [Batch]
        # State value V(s_t): [Batch, 1]

        # Detach hidden states from the VLA backbone's computation graph
        h = hidden_states.detach()
        
        # 1. Compute attention scores and normalize
        # e shape: [Batch, Seq_Len, 1]
        e = self.attn_proj(h)                  
        # alpha shape: [Batch, Seq_Len, 1]
        alpha = F.softmax(e, dim=1)            
        
        # 2. Compute pooled representation z_pool
        # z_pool shape: [Batch, Hidden_Dim]
        z_pool = torch.sum(alpha * h, dim=1)   
        
        # 3. Incorporate step embedding
        # e_step shape: [Batch, Hidden_Dim]
        e_step = self.step_emb(step_t)         
        
        # 4. Final state value estimation
        # v shape: [Batch, 1]
        v = self.mlp(z_pool + e_step)          
        
        return v
\end{lstlisting}

\subsection{Token-level Policy Optimization}
Defining the granularity of the optimization objective is critical for training stability, leading us to distinguish between three primary approaches: chunk-level, action-level, and our adopted token-level optimization. 

Standard chunk-level methods treat the entire sequence of $K$ tokens as a single unified action. The joint probability is calculated as $P(a_{1:K}|o) = \prod_{k=1}^K \pi(a_k | o, a_{<k})$, which frequently leads to severe numerical instability. Because the joint probability is a product of multiple values between 0 and 1, the resulting magnitude can become infinitesimally small, causing the PPO importance sampling ratio $\rho_t(\theta) = \frac{\pi_\theta(a_{1:K}|o)}{\pi_{\theta_{old}}(a_{1:K}|o)}$ to fluctuate wildly. Such fluctuations often trigger the PPO clipping mechanism, leading to a total loss of gradient signals and stalled training.

Alternatively, an action-level approach might compute the loss based on the mean or aggregated probability of the tokens within a single action dimension. However, this still risks blurring the independent learning signals of individual tokens, failing to provide precise credit assignment across the multi-dimensional action space of the VLA.


In contrast, we implement a token-level optimization strategy. By treating each action token as an independent decision point conditioned on the shared observation, the importance sampling ratio is redefined for each individual token $k$:
\begin{equation}
\rho_{t,k}(\theta) = \frac{\pi_\theta(a_{t,k} \mid o_t)}{\pi_{\theta_{old}}(a_{t,k} \mid o_t)}
\end{equation}
By calculating the clip loss at the token level, we prevent outlier tokens from causing the entire multi-dimensional action's gradient to be clipped, thereby retaining a much higher density of training signals. This approach ensures numerical robustness by avoiding the vanishing product problem and provides precise credit assignment across specific components of the action space.


\subsection{Dynamic Weighted Resampling}
To improve sample efficiency and balance the learning progress across diverse tasks, we implement a Dynamic Weighted Resampling (DWR) mechanism. DWR adaptively adjusts the sampling probability of each task based on its recent performance. 

Instead of relying on a static uniform distribution, we maintain a sliding window of recent episode outcomes for each task. The sampling weight is dynamically computed proportional to the recent failure rate, complemented by a Laplace smoothing factor:

\begin{lstlisting}[language=Python, basicstyle=\ttfamily\small, keywordstyle=\color{blue}, commentstyle=\color{gray}, frame=single, breaklines=true]
import numpy as np

class DynamicWeightedResampler:
    def __init__(self, num_tasks, window_size=100, eps=1.0):
        self.num_tasks = num_tasks
        self.window_size = window_size
        self.eps = eps
        
        # Circular buffer for history (1.0 for success, 0.0 for failure)
        # Initialized to ones to prevent early bias against unattempted tasks
        self.history = np.ones((num_tasks, window_size))
        self.ptr = 0

    def update_history(self, task_idx, success_flag):
        self.history[task_idx, self.ptr] = success_flag
        self.ptr = (self.ptr + 1) % self.window_size

    def sample_task(self):
        # 1. Count failures (Total window size - successes)
        success_counts = np.sum(self.history, axis=1)
        failure_counts = self.window_size - success_counts
        
        # 2. Apply Laplace smoothing to compute sampling weights
        weights = failure_counts + self.eps
        
        # 3. Normalize to obtain sampling probabilities
        probabilities = weights / np.sum(weights)
        
        # 4. Sample task index based on the generated distribution
        return np.random.choice(self.num_tasks, p=probabilities)
\end{lstlisting}

This DWR strategy effectively directs computational resources toward challenging tasks where performance lags. Furthermore, the smoothing term \texttt{eps} guarantees that all tasks maintain a non-zero sampling probability, effectively preventing catastrophic forgetting of previously mastered skills while the agent tackles new challenges.

\subsection{Asynchronous Parallel Data Prefetching}
To address the I/O bottlenecks inherent in experience replay sampling and trajectory processing, we implement a data prefetching architecture based on asynchronous parallel logic. By maintaining a background asynchronous loop within the trainer that operates independently of the primary training task, AcceRL achieves an effective decoupling of data preparation from gradient updates. Functioning as a producer, this background process continuously monitors the distributed replay buffer and triggers cross node trajectory sampling once the required threshold is met. This concurrency ensures that while the GPU executes gradient computations for the current batch, system resources are simultaneously preparing the data for the subsequent iteration. AcceRL simplifies the training critical path by delegating observation tensorization and batching operations to the background prefetcher. The trainer extracts ready to use super-batches directly from a local cache. This streamlined data loading mechanism effectively significantly reduces I/O latency, maximizing hardware utilization and accelerating convergence in complex embodied tasks.

\subsection{High-Performance NCCL-based Weight Synchronization}
The performance of an online RL framework highly depends on the efficiency of policy synchronization between the trainer and the Inference Worker. If the Inference Worker uses stale weights due to synchronization delays, it will cause severe policy staleness, which may disrupt the stability and sample efficiency of near-on-policy training.

To address this, our framework uses a weight synchronization mechanism based on NCCL collective communication, which upgrades the synchronization path from traditional CPU/TCP or shared storage to the level of GPU-to-GPU collective communication:
\begin{enumerate}
    \item \textbf{Intra-node Synchronization Group Construction}: In a multi-node distributed deployment, the synchronization is organized by group. The trainer-side source and the Inference Worker form an independent NCCL Broadcast Group, ensuring that weight synchronization is handled locally. It fully utilizes the high bandwidth of NVLink and avoids the potential bottlenecks of the cross-node InfiniBand network.
    \item \textbf{In-place Tensor Update}: Whenever the training side completes a gradient update, the trainer directly broadcasts the model's trainable parameters and necessary buffers in a fixed sequence to the inference GPU. After the inference GPU receives the tensors, it directly overwrites the inference model weights through in-place operations. This avoids the overhead of Python serialization, host-side memory copying, or object storage distribution.
\end{enumerate}

To further reduce trainer waiting latency at synchronization points, we propose a lightweight Inference Drain mechanism. The trainer sends a drain signal to Inference Worker well before parameter updates finish. Upon receiving this signal, Inferences stop scheduling new forward batches. Instead, they first complete all ongoing computations, then switch to the state for receiving updated weights. This mechanism effectively prevents the trainer from being blocked by the long-duration tail of inference forward computations, while also guaranteeing the atomicity and version consistency of model updates on the inference side. Through this fine-grained synchronization control, the system significantly enhances overall hardware coordination efficiency while maintaining high-frequency policy updates.